\documentclass[11pt,a4paper]{article}
\usepackage[hyperref]{emnlp2020}
\usepackage{times}
\usepackage{latexsym}

\usepackage{microtype}

\aclfinalcopy 

\usepackage{hyperref}
\usepackage{marvosym}
\usepackage{amsmath}
\usepackage{amsthm}
\usepackage{mathtools}
\usepackage{verbatim}
\usepackage{booktabs}
\usepackage{ifthen}
\usepackage{booktabs}
\usepackage{blindtext}
\usepackage{amssymb}
\usepackage{xcolor}
\usepackage{nicefrac}
\usepackage{enumitem}
\usepackage{tikz}
\usetikzlibrary{positioning}
\usetikzlibrary{calc,through,backgrounds}

\usepackage{marvosym}

\newcommand\simplex{\triangle}
\DeclareMathOperator*{\softmax}{\mathsf{softmax}}
\DeclareMathOperator*{\sparsemax}{\mathsf{sparsemax}}
\DeclareMathOperator*{\argmax}{\mathsf{argmax}}
\DeclareMathOperator*{\JS}{\mathsf{JS}}
\DeclareMathOperator*{\KL}{\mathsf{KL}}
\DeclareMathOperator*{\epsilonppl}{\epsilon-\mathsf{ppl}}
\newcommand{\bs}[1]{\boldsymbol{#1}}

\newcommand{\andre}[1]{{\textcolor{blue}{\bf [{\sc Andre:} #1]}}}

\definecolor{cadmiumgreen}{rgb}{0.0, 0.42, 0.24}

\title{Sparse Text Generation}

\author{Pedro Henrique Martins\textsuperscript{\Neptune} \quad
        Zita Marinho\textsuperscript{\Moon\Scorpio} \and
        Andr\'e F.~T. Martins\textsuperscript{\Neptune\Pluto\Saturn} \\
\textsuperscript{\Neptune}Instituto de Telecomunica\c{c}\~oes~
\textsuperscript{\Moon}Priberam Labs~
\textsuperscript{\Scorpio}Institute of Systems and Robotics\\
\textsuperscript{\Pluto}LUMLIS (Lisbon ELLIS Unit), Instituto Superior T\'ecnico~
\textsuperscript{\Saturn}Unbabel\\
Lisbon, Portugal\\
\href{mailto:pedrohenriqueamartins@tecnico.ulisboa.pt}{\tt pedrohenriqueamartins@tecnico.ulisboa.pt},\\
\href{mailto:zita.marinho@priberam.pt}{\tt zita.marinho@priberam.pt}, \quad
\href{mailto:andre.martins@unbabel.com}{\tt andre.t.martins@tecnico.ulisboa.pt}.
}

\date{}

\begin{document}
\maketitle

\begin{abstract}
Current state-of-the-art text generators build on powerful language models such as {GPT-2}, achieving impressive performance. However, to avoid degenerate text, they require sampling from a modified softmax, via temperature parameters or ad-hoc truncation techniques, as in top-$k$ or nucleus sampling. This creates a mismatch between training and testing conditions. 
In this paper, we use the recently introduced entmax transformation  
to train and sample from a natively sparse language model, avoiding this mismatch.
The result is a text generator with favorable performance in terms of fluency and consistency, fewer repetitions, and n-gram diversity closer to human text. 
In order to evaluate our model, we propose three new  metrics for comparing sparse or truncated distributions: $\epsilon$-perplexity, sparsemax score, and Jensen-Shannon divergence. 
Human-evaluated experiments in story completion and dialogue generation show that
entmax sampling leads to more engaging and coherent stories and conversations.
\end{abstract}

\section{Introduction}

\begin{table}[ht]
\vspace{\baselineskip}
\centering \small
\setlength{\tabcolsep}{0.5ex}
\begin{tabular}{lp{0.82\columnwidth}}
\toprule
\textbf{Context:} &  {\it ``Are they ready to go live?" Zuko asked. ``Yup, in just a minute" Tomi replied as he leaned back in his reclining chair. This was a moment that the entire galaxy would see. After 10 years...}
\\
\midrule
\textbf{Greedy:}  & ... of study, the scientists had finally come to the conclusion that the universe was not a place for living. \textcolor{red}{\it The scientists} had been studying \textcolor{red}{\it the universe} for over a century.  \textcolor{red}{\it They had been studying the universe for over a century.}\vspace{0.2cm}\\ 
\textbf{Top-$k$:} & ... of studying this phenomenon, the team had finally come up with a plan to \textcolor{red}{\it study} what the aliens were doing. \textcolor{red}{\it The team} would be looking for a way to create a space station. ``And what would you call it?" \textcolor{red}{\it ``What would you call the space station?"}\vspace{0.2cm} \\
\textbf{Nucleus:} & ... of diligent testing, \textcolor{red}{\it and some test run after that}, the results were well received. ``they're waiting for you, because \textcolor{red}{\it there's no way they would.}'' \textcolor{red}{\it Stowedly at  a  local wedding  place, the two archaeologists spoke together.}\vspace{0.2cm} \\
\textbf{Entmax:} & ... of hibernation, Tomi was back. These last days had been a  significant step forward in  his mission. This time, Tomi was not alone. All  the empires had aligned together and the world's leadership began to openly support his mission. \\
\bottomrule
\end{tabular}
\caption{Completion of a story from the WritingPrompts dataset, using the greedy decoding, top-$k$ sampling, nucleus sampling, and entmax sampling (our proposal) methods. Repetitions and off-topic text are \textcolor{red}{\it highlighted.}\label{tab:examples}}
\label{example}
\end{table}

The last few years have seen a tremendous progress in neural language modeling, with a strong impact on key NLP tasks such as machine translation \citep{sutskever2014sequence,Bahdanau2015,vaswani2017attention}, speech recognition \citep{chorowski2015attention}, summarization \citep{rush2015neural}, and open-ended text generation \citep{radford2019language}. 
Benefiting from large amounts of data, models such as 
Transformer-XL \cite{dai2019transformer} and  GPT-2 \cite{radford2019language} 
have achieved impressive perplexity scores on language modeling. 
However, the generated text is still often repetitive and incoherent (Table~\ref{tab:examples}). 

A downside of current approaches is the mismatch between training and testing conditions: models are usually trained to maximize the likelihood of observed text. 
However, when generating, state-of-the-art models sample from a {\bf truncated} and {\bf renormalized} softmax distribution \citep{fan2018hierarchical,holtzman2020curious}. 
They do so as a compromise to avoid two extremes: 
a deterministic search for the most probable sentence (via greedy decoding or beam search) usually results in dull and  repetitive ``degenerate text'' 
\citep{lidiversity,li2017-adversarial,holtzman2020curious}; stochastically sampling from the full softmax distribution, on the other hand, often generates many implausible words from the tail of the distribution \citep{fan2018hierarchical}.   
The recently proposed \emph {nucleus sampling} approach 
\citep{holtzman2020curious} sets the truncation point based on the cumulative distribution function, \emph{i.e.}, it considers the top words with a cumulative probability $P$. 
In this approach  
the number of words to sample from are selected according to the context, in opposition to 
\emph {top-$k$ sampling}~\citep{fan2018hierarchical}, which samples from the $k$ most probable words. 
However, the ``sparsity'' introduced by both methods is artificially imposed at run time, not learned during training. 

A second problem is that it is hard to compare different truncation strategies---for example, we cannot easily evaluate how the resulting truncated distributions behave as language models, since the most widely used metric for language modeling---perplexity---cannot handle sparse distributions: if a model assigns zero probability to a single reference word, it gets infinite perplexity for the entire sample. For this reason, previous works  generate from a  {\it truncated} softmax, but  report the perplexity of the {\it full} softmax distribution \citep{welleck2020neural, li2019don}. Others use the latter to compare  perplexity on the generated text with that on human text \citep[\S4.2]{holtzman2020curious}, or resort to distributional statistics \citep{zhu2018texygen}.

In this paper,  
we propose a new approach---{\bf entmax sampling} (\S\ref{sec:entmax})---that eliminates the mismatch between training and test conditions. 
Key to our approach is the recently proposed entmax transformation \citep{peters2019sparse}.  
Entmax transforms a vector of scores into a {\bf sparse probability distribution}, preventing implausible words from receiving {\it any} probability mass. 
Moreover, it does so {\bf natively}: it comes with a well-defined loss function that allows it to learn its sparsity automatically from the data, during training. 
This results in a new stochastic text generator where the number of possible word types varies with the context (like nucleus sampling), but that generates by sampling directly from its output distribution (like softmax),
and where the sparsity of this distribution is present during training (unlike any existing method).

As a second contribution, we propose {\bf three new metrics}
to support the evaluation of sparse language models
 (\S\ref{sec:metrics}): $\epsilon$-perplexity, sparsemax score, and Jensen-Shannon divergence.
We show that these metrics are well supported theoretically 
and can be used 
to compare our method with various truncation and temperature techniques.  

Experiments in {language modeling}, {story completion}, and {dialogue generation} (\S\ref{sec:experiments}) show that entmax sampling generates more diverse text and fewer repetitions than nucleus and top-$k$ sampling.\footnote{The code used for the experiments and for the proposed metrics is available at \url{https://github.com/deep-spin/sparse_text_generation}.}

\subsection{Related work}

\paragraph{Decoding methods.} While greedy decoding and beam search are popular strategies for sequence-to-sequence tasks, such as machine translation, \citet{knowles2016analyzing} and \citet{stahlberg2019nmt} showed that searching for the most probable sentence in a model trained with likelihood maximization has a bias for short sentences. 
In open-ended generation, 
\citet{fan2018hierarchical} and \citet{holtzman2018learning,holtzman2020curious}
have shown that these methods lead to repetitions and dull text. 
To overcome this, several authors proposed beam search variants which promote word diversity \citep{li2016simple,vijayakumar2018diverse,kulikov2018importance}. 
An alternative to deterministic text generation is to sample directly from the softmax distribution. However, since the probability mass tends to accumulate in a long tail, this procedure  generates  unlikely words too often, leading to degenerate text \citep{fan2018hierarchical,holtzman2020curious}. This can be mitigated by lowering the softmax temperature \citep{ficler2017controlling}, 
by sampling from the top-$k$ most probable words only
\citep{fan2018hierarchical, radford2019language}, or through {nucleus sampling} \citep{holtzman2020curious}. 
We compare against these methods in \S\ref{sec:experiments}.

\paragraph{Diversity-promoting models.} In addition to new decoding methods, models that aim to increase word diversity and diminish repetition have also been introduced. \citet{xu2018diversity} proposed a diversity-promoting generative adversarial network, which rewards
novel and fluent text. \citet{holtzman2018learning} proposed augmenting the language model with several discriminators. More recently, \citet{welleck2020neural} proposed augmenting the loss with an unlikelihood term that penalizes the generation of tokens that are present in the context, 
a method against which we compare in \S\ref{sec:experiments}.

\paragraph{Sparse transformations and losses. } 
At the core of our work are sparse alternatives to the softmax transformation. 
\citet{martins2016softmax} proposed sparsemax and applied it to multi-label classification. 
This was generalized by \citet{peters2019sparse} via their $\alpha$-entmax transformation, which was applied to sequence-to-sequence models for morphological inflection and machine translation. In contrast to our work, they performed {\it deterministic} decoding with beam search, and they did not consider open-ended generation.

\paragraph{Evaluation metrics.}
The most common metrics to evaluate text generation models are perplexity \citep{jelinek1977perplexity} and BLEU \citep{papineni2002bleu}. For open generation,
\citet{zhu2018texygen} observed that ``no single metric is comprehensive enough''.   Other evaluations include corpus $n$-gram overlap \citep{yu2017seqgan,press2017language}, and the Fr\'{e}chet distance \citep{cifka2018eval}. 
These approaches are aimed at the (harder) problem of evaluating the quality of generated text. By contrast, our paper proposes new metrics for evaluating {\it language models} in the task of predicting the next word conditioned on ground truth context (like perplexity does), but supporting sparse probability distributions (which perplexity does not).

\section{Language Modeling}\label{sec:lm}

Language models assign probability to word sequences $x = \langle \textsc{start},x_1,\dots,x_T,\textsc{stop} \rangle$, where each $x_t$ is in a vocabulary $\mathcal{V}$, and $T \in \mathbb{N}$. 
This probability can be written as $p_\theta(x) = \prod_{t=1}^{T+1} p_\theta(x_t\mid x_{<t})$. We would like the model $\theta$ to assign high probability to real sentences, i.e., each distribution $p_\theta(\cdot \mid x_{<t})$ should assign a large probability value to the ground truth $x_t$. 

Given a set $\mathcal{S}$ of training sentences, the usual strategy for learning the language model parameters $\theta$ is to minimize the negative log-likelihood:
\begin{equation}\label{eq:nll}
    \mathcal{L} (\theta) = -\sum_{i=1}^{|\mathcal{S}|} \sum_{t=1}^{T_i} \log p_\theta (x_t^i|x_{<t}^i).
\end{equation}
The standard choice to model $p_\theta (\cdot|x_{<t})$ in Eq.~\ref{eq:nll} is to  compute a score vector $\bs{z}_t$ by conditioning on the context $x_{<t}$, and then applying a softmax transformation, $p_\theta (\cdot|x_{<t}) = \softmax(\bs{z}_t)$, where 
\begin{equation}
    [\softmax(\bs{z}_t)]_k =\dfrac{\exp({z_t}_k)}{\sum_j \exp({z_t}_j) }.    
\end{equation}

At decoding time, the language model generates sentences one word at a time, by sampling from the learned probability distribution. 
However, softmax yields a {\bf dense} distribution, \emph{i.e.}, some probability mass (even if small) is assigned to all the words in the vocabulary. 
\citet[\S3]{holtzman2020curious} have shown that, if we sample from this distribution directly, the resulting text becomes 
degenerate, with common incoherences arising due to the unreliability of the tail of the distribution. 
This motivated a line of work proposing ``ad-hoc'' modifications to the softmax distribution, to reduce the effect of the tail. 
Two of the most successful techniques, top-$k$ and nucleus sampling \citep{fan2018hierarchical, holtzman2020curious}, 
do so by truncating and renormalizing the distribution $p_\theta (\cdot|x_{<t})$. Note that these techniques are applied only at decoding time---during training the original softmax distribution is left untouched, being used as part of the optimization of the cross-entropy loss. 

Our alternative to these ad-hoc modifications builds on {\it learnable} sparse transformations, as we shall see in \S\ref{sec:entmax}. 
These transformations can produce  sparse, zero-tailed probability distributions, learning the amount of sparsity from data. Therefore, sampling from these distributions directly is a natural way to prevent degenerate text.

\section{Entmax Sampling}\label{sec:entmax}

Key to our method is the recently proposed $\alpha$-entmax family of transformations\footnote{\url{https://github.com/deep-spin/entmax}.} 
\citep{peters2019sparse}, parametrized by a scalar parameter $\alpha \ge 1$:
\begin{equation}
    \alpha \text{-} \mathsf{entmax}(\bs{z}_t) :=  \argmax_{\bs{p}\in \simplex^d} \bs{p}^\top \bs{z}_t + \mathsf{H}_\alpha(\bs{p}).
\end{equation}
Above, $\textstyle \simplex^d \coloneqq \left\{ \bs{p} \in \mathbb{R}^{d} \mid \sum^d_{i=1} p_i=1, \bs{p}\geq\bs{0}\right\}$ is the probability simplex, and $\mathsf{H}_{\alpha}$ is the Tsallis $\alpha$-entropy \citep{tsallis1988possible}:
\begin{equation}
    \mathsf{H}_\alpha (\bs{p}) := \begin{cases}
                                                \frac{1}{\alpha(\alpha-1)}\sum_j(p_j-p_j^\alpha), &  \alpha \neq 1\\
                                                -\sum_j p_j \log p_j, & \alpha=1.
                                            \end{cases}
\end{equation}
With $\alpha=1$ and $\alpha=2$, 
we recover the Shannon and Gini entropies, respectively.\footnote{\label{foot:gini}The Gini entropy is 
$\mathsf{H}_2(\bs{p}):=\frac{1}{2}\sum_j p_j(1-p_j)$.} 
When $\alpha \rightarrow \infty$, $\mathsf{H}_{\alpha}(\bs{p}) \rightarrow 0$. 
Thus, $\mathsf{1}\text{-}\mathsf{entmax}$, $\mathsf{2}\text{-}\mathsf{entmax}$, and $\mathsf{\infty}\text{-}\mathsf{entmax}$ recover $\softmax$, $\sparsemax$, and $\argmax$, respectively. 
\citet{blondel2019learning} have shown that, for $\alpha>1$, entmax is able to output {\bf sparse} probability distributions, where some words get {\bf exactly} zero probability, whereas softmax ($\alpha = 1$) does not have this capability. 

\paragraph{How can we learn this output sparsity during training?}
Following \citet{peters2019sparse}, we 
replace the negative log-likelihood loss in Eq.~\ref{eq:nll} by 
\begin{equation}
    \mathcal{L}(\theta) = \sum_{i=1}^{|\mathcal{S}|} \sum_{t=1}^{T_i} \ell_\alpha(\bs{z}_t(\theta, x_{<t}), x_t),
\end{equation}
where $\ell_\alpha(\bs{z}_t, x)$ is 
the {\bf $\alpha$-{entmax} loss}:
\begin{equation}\label{eq:entmax_loss}
    \ell_\alpha(\bs{z}_t, x) := (\bs p_\theta-\bs{e}_{x})^\top \bs{z}_t + \mathsf{H}_\alpha(\bs p_\theta),
\end{equation}
where $\bs p_\theta=\alpha \text{-} \mathsf{entmax}(\bs{z}_t)$, and $\bs{e}_{x}$ is the  one-hot vector corresponding to the ground truth word $x$. 
When $\alpha=1$, we still recover the negative log-likelihood, 
$\ell_\alpha(\bs{z}_t, x) = -\log p_\theta(x)$, 
and, when $\alpha=2$, this corresponds to the {\bf sparsemax loss} \citep{martins2016softmax},  to be revisited in \S\ref{sp_score}. 

Entmax losses belong to the wider class of Fenchel-Young losses \citep{blondel2019learning} and, consequently, are convex on $\bs{z}$ and differentiable (with gradient $\nabla_{\bs{z}} \ell_\alpha(\bs{z}, x) = -\bs{e}_{x} + \bs p_\theta$). 
For $\alpha>1$, they have a {\bf separation margin}: the loss is zero iff 
$z_{tx} \ge z_{tx'} + \frac{1}{\alpha-1}$ for all $x' \ne x$, in which case $\bs p_\theta = \bs{e}_{x}$, i.e., the model puts all its probability mass in the correct word. 
This allows the model to be adaptive to the degree of uncertainty present: in some cases there are few plausible words, so most words should have probability zero, while in other cases a higher number of words are plausible and should be given probability mass.

\paragraph{Entmax sampling.} 
At test time, we simply sample from the categorical distribution obtained by applying the entmax transformation to the scores $\bs{z}_t$ given by the model:
\begin{align}
    x_t \sim p_\theta(\cdot \mid x_{<t}) = \alpha\text{-}\mathsf{entmax}(\bs{z}_t(\theta, x_{<t})).
\end{align}
Note that, 
in contrast to previously proposed methods such as top-$k$ sampling and nucleus sampling \citep{holtzman2020curious}, we sample directly from the learned sparse probability distribution over the words, without any calibration or ad-hoc modification. 
As in nucleus sampling and in opposition to top-$k$ sampling, entmax sampling considers a varying number of tokens depending on the context. Moreover, as we show  
in Table~\ref{word_statistics}, with entmax sampling this variability is higher. 

\section{Evaluation Metrics}\label{sec:metrics}

Language models are commonly evaluated by computing their perplexity ($\mathsf{ppl}$) on held-out data. Perplexity assesses the ability of a language model to predict the next word given the context:
\begin{equation}\label{eq:ppl}
    \mathsf{ppl} = \mathsf{exp}\left({-\frac{1}{T}\sum_{t=1}^T  \log p_\theta(x_t \mid x_{<t})}\right).
\end{equation}
However, its computation involves the logarithm of a probability. 
This poses a problem when we are using sparse or truncated probability distributions, since $\lim_{p\rightarrow 0}\log p = -\infty$. 
Usually, authors report the values for perplexity computed on the original probability distribution, before truncation. However, this metric does not allow different sparse decoding strategies to be compared.\footnote{This is important not only when we have sparse or truncated probability distributions, but also to compare language models using different vocabularies: when using perplexity, if the ground truth word is not contained in the vocabulary, one usually considers the probability attributed to an {\sc unk} token instead of a zero probability, which leads to an unfair comparison between models with different vocabularies.} 
As an alternative, we propose 
three different metrics (to better understand these metrics, comparative plots are shown in Fig.~\ref{fig:plot_ppl}, App.~\ref{sec:plots_metrics}).

\paragraph{$\epsilon$-perplexity.} To be able to compute the perplexity for sparse distributions, the simplest approach 
is to smooth it by adding a small value $\epsilon$ to all terms followed by renormalization, as in additive (Laplace) smoothing \citep{chen1999empirical}:
\begin{equation}
    \epsilonppl = \mathsf{exp}\left(-\frac{1}{T}\sum_{t=1}^T  \log \dfrac{p_\theta(x_t \mid x_{<t})+\epsilon}{1 + \epsilon|\mathcal{V}|}\right).
\end{equation}
Note that, like perplexity, $\epsilonppl$ only depends on $\theta$ via the probabilities assigned to the reference words. 
When used as a metric for a language model, we may regard $\epsilon$ as a calibration parameter that the language model is allowed to tune to better match the reference. 
We show in App. \ref{epsilon_opt} that 
the optimal value of $\epsilon$ (i.e., the one that leads to the smallest $\epsilonppl$) can be obtained from these probabilities by solving a simple convex optimization problem---this is convenient, since it avoids the need for manual tuning. 
A disadvantage of $\epsilonppl$ is that it still does not evaluate the original sparse distribution, but rather a modified version of it.
However, when applied to variants of truncated softmax, by collapsing all the truncated probabilities to the same value $\epsilon$, it is useful to measure how much truncation deteriorates its ability to rank words, compared to softmax.

\paragraph{Sparsemax score.}
\label{sp_score}
We can derive a more interesting metric that handles sparse distributions directly. 
By setting $\alpha=2$ in 
Eq.~\ref{eq:entmax_loss},
\footnote{If we set $\alpha=1$ instead, we revert to perplexity.} 
we obtain the sparsemax loss proposed by \citet{martins2016softmax}, $\ell_2(\bs{z}, x) = (\bs{p}_\theta-\bs{e}_{x})^\top \bs{z} + \mathsf{H}_2(\bs{p}_\theta)$. 
We define the {\bf sparsemax score} ($\mathsf{sp}$) as:
\begin{align}\label{eq:sparsemax_score}
    \mathsf{sp} &= 1 - \min \{\ell_2(\bs{z}, x) \mid \mathsf{sparsemax}(\bs{z}) = \bs{p}_\theta\} \nonumber\\
    &= 1 - (\bs{p}_\theta-\bs{e}_{x})^\top \bs{p}_\theta - \mathsf{H}_2(\bs{p}_\theta)\nonumber\\
    &= {p}_\theta(x) + \mathsf{H}_2(\bs{p}_\theta),
\end{align}
where $\mathsf{H}_2$ is the Gini entropy (see footnote \ref{foot:gini}). 
Unlike perplexity, this score is bounded. In fact, it is always between 0 (when $\bs{p}_\theta = \bs{e}_{x'}$ with $x' \ne x$) and 1 (when $\bs{p}_\theta = \bs{e}_{x}$). 
We prove this fact in App.~\ref{sec:proof_sparsemax_bounded}. 
Interestingly, when the model $\bs{p}_\theta$ is deterministic (e.g., when it comes from greedy search), we have $\mathsf{H}_2(\bs{p}) = 0$, and the sparsemax score simply becomes the {\bf word accuracy}. 
In the opposite case, when $\bs{p}_\theta$ is uniform, we obtain 
$\mathsf{sp} = \frac{1}{|\mathcal{V}|} + \frac{1}{2}\left(1 - \frac{1}{|\mathcal{V}|}\right) \rightarrow 0.5$ when 
$|\mathcal{V}|\rightarrow \infty$. 

We show in App.~\ref{sec:patrick_fischer} that this score is related to the {Patrick-Fischer distance} \citep[p.~262]{patrick1969nonparametric,deza2009encyclopedia}.

\paragraph{Jensen-Shannon Divergence.}
\label{js}
Given two discrete probability distributions $\bs{p}_\theta$ and $\bs{q}$, and denoting their mixture (arithmetic mean) as $\bs{m} := \frac{\bs{p}_\theta+\bs{q}}{2}$, and the Kullback-Leibler divergence as $\KL$, the Jensen-Shannon divergence is defined as: 
\begin{align}
    \JS(\bs{p}_\theta,\bs{q}) &= \frac{1}{2}\,\KL(\bs{p}_\theta||\bs{m}) + \frac{1}{2}\,\KL(\bs{q}||\bs{m})\nonumber\\ 
                       &= \frac{1}{2}\sum_{x \in \mathcal{V}}  p_\theta(x) \log \left(\dfrac{p_\theta(x)}{m(x)}\right)\nonumber\\ 
                       &+ \frac{1}{2}\sum_{x \in \mathcal{V}} q(x) \log \left(\dfrac{q(x)}{m(x)}\right).
\end{align}

The Jensen-Shannon divergence can be interpreted as a mutual information as follows \citep{grosse2002analysis,banerjee2005clustering}: 
consider a two-step process where we first toss a fair coin $B \sim \mathrm{Bernoulli}(\frac{1}{2})$. If the outcome is heads, we sample the next word $X$ according to the model $p_{\theta}(\cdot)$; if it is tails, we sample $x\sim q(\cdot)$. A word generated according to this process is governed by the mixture $m(\cdot)$, $x \sim m(\cdot)$. The Jensen-Shannon divergence between $\bs{p}_{\theta}$ and $\bs{q}$ is the mutual information between the random variables $B$ and $X$, which equals $\mathsf{H}(B) - \mathsf{H}(B \mid X)$, where $\mathsf{H}$ is the Shannon entropy and $\mathsf{H}(B \mid X) = \sum_{x \in \mathcal{V}} m(x) H(B \mid X=x)$ is the conditional entropy. 
Hence, the Jensen-Shannon divergence can be seen as the reduction of uncertainty about the source $B$ when we observe a sample $x$ from the mixture $m(\cdot)$. The more similar the two distributions $\bs{p}_{\theta}$ and $\bs{q}$ are, the smaller this reduction is.

In our experiments, we report the 
$\mathsf{JS}$ as an evaluation metric for language models, setting $\bs{q} = \bs{e}_x$ (i.e., a one-hot distribution placed on the ground truth word $x$) and averaging the $\mathsf{JS}$ over the words. 
Like the sparsemax score described above, 
the 
$\mathsf{JS}$ is bounded: it is zero if $\bs{p}_\theta = \bs{e}_x$, and maximal ($\log(2)$) when $\bs{p}_\theta$ is a one-hot distribution placed on a different word. We show in App.~\ref{js_} that, like  $\epsilonppl$ (but unlike $\mathsf{sp}$), the $\mathsf{JS}$ only depends on $\theta$ via the probabilities assigned to the reference words.

\paragraph{Comparing multiple models.}
The generalized JS allows to compare two or more trained models:
\begin{align}
    \JS(\bs{p}^1,\dots,\bs{p}^K) = \frac{1}{K}\sum_{k=1}^K \KL(\bs{p}^k \| \bs{m})
\end{align}
where $\bs{p}^1,\dots,\bs{p}^K$ are the probability distributions of the different models and $\bs{m} = \frac{1}{K}\sum_{k=1}^K \bs{p}^k$ is their mixture. This property can be useful for measuring the diversity between multiple models (e.g., when used in an ensemble system). We use this metric in App.~\ref{js_comp_models} to rank the sentences in which the different models we compare disagree the most. 

\section{Experiments}\label{sec:experiments}

We compare the different methods 
in three NLP tasks: language modeling (\S\ref{lm_exp}), story completion (\S\ref{sc_exp}), and dialogue generation (\S\ref{rp_exp}). In language modeling, we evaluate the model's 
fluency, while in story completion we also evaluate if the methods generate coherent and ``interesting'' text. In dialogue generation, we evaluate the methods' performance in an interactive task.

\subsection{Language Modeling}
\label{lm_exp}
\begin{table*}[t]
\vspace{\baselineskip}
\centering \small
\setlength{\tabcolsep}{0.7ex}
\resizebox{\textwidth}{!}{
\begin{tabular}{l@{\hspace{0.5ex}}rrrrr@{\hspace{2.5ex}}rrrrr@{\hspace{2.5ex}}rrrrr}
\toprule
& \multicolumn{5}{c}{WikiText-2} & \multicolumn{5}{c}{WikiText-103}& \multicolumn{5}{c}{BookCorpus}\\
\midrule
     & \textsf{sp} & \textsf{JS} & $\epsilonppl$ & \textsc{rep} & \textsc{wrep} & \textsf{sp} & \textsf{JS} & $\epsilonppl$ & \textsc{rep} & \textsc{wrep} & \textsf{sp} &\textsf{JS} & $\epsilonppl$ & \textsc{rep} & \textsc{wrep} \\
\midrule
Softmax & .682 & .376 & \textbf{12.74} & \textbf{.407} & .174 & .683 &  .375 & 13.29 & .349 & .162 & .680 & .366 & 10.80 &  .376 & .183  \\
Softmax-$\tau$ & .680 & .369 & 12.97 & .414 & .176 & .682 & .368 & 13.65 & .359 & .168 & .677  & .363 & 10.96 & .391 & .191  \\
Greedy & .491 & \textbf{.358} & 459.13 & .525 & .232 & .499 &  \textbf{.355} & 512.50 & .450 & .210 & .489 & \textbf{.354} & 506.86 & .461 & .211 \\
Top-$k$ & .682 & .363	& 20.93 & .437 & .196 & .683 & .364 & 21.90  & .373 & .181 & .676 & .360 & 22.25 & .399 & .203 \\
Nucleus & .684 & .371 & 14.65 & .412 & .175 & .686 & .370 & 15.51 & .357 & .167 & .678 & .362 & 16.48 &  .392 & .193 \\
Unlikelihood & .473 & .365 & 599.65 & .467 & .210 & .471 & .366 & 610.06 & .410 & .200 & .475 & .364 & 587.04 & .418 & .203  \\
Entmax & \textbf{.688} & .369 & 13.91 & \textbf{.407} & \textbf{.171} & \textbf{.694} & .373 & \textbf{13.23} & \textbf{.346} & \textbf{.160}  & \textbf{.687} & .362 & \textbf{10.70} & \textbf{.374} & \textbf{.179}\\
\bottomrule
\end{tabular}
}
\caption{Language model evaluation on WikiText-2, WikiText-103, and BookCorpus test sets. For all metrics except \textsf{sp}, lower is better. See App.~\ref{finetuning_lm} for the results on the validation set.}
\label{results_lm_auto}
\end{table*}

\paragraph{Datasets and metrics.}
We performed experiments on three widely used language modeling datasets: WikiText-2 and WikiText-103 \cite{merity2016pointer}, and BookCorpus \cite{Zhu2015AligningBA}. 
WikiText-2 and WikiText-103 are composed of Wikipedia articles, 
comprising around 2 and 100 million tokens for training, respectively. Their validation and test sets have 
217,000 and 245,000 tokens. 
BookCorpus is composed of 11,038 freely available books. 
We used the standard split: 800 million tokens for training, 260,000 for validation, and 280,000 for testing. 

We report the sparsemax score, Jensen-Shannon, and $\epsilon$-perplexity 
(\S\ref{sec:metrics}) to evaluate the methods' fluency, and the \textsc{rep} and \textsc{wrep}\footnote{\textsc{rep} measures the number of times that a word from the previous $l$ words is repeated, when generating the following word. \textsc{wrep} does the same, discarding words that are also repeated in the ground truth. We report the average of \textsc{rep} and \textsc{wrep} for $l\in\{16,32,128,512\}$.} \citep{welleck2020neural} to evaluate the methods' tendency to generate repetitions. All metrics are computed at the BPE level~\cite{sennrich2016neural}.

\paragraph{Fine-tuning GPT-2.}
We fine-tuned the GPT-2 medium model  \cite{radford2019language}, which consists of a 24 layer transformer with 345 million parameters.\footnote{We use the PyTorch re-implementation at \url{https://github.com/huggingface/transformers}.} 
We fine-tuned three models with the following losses: negative log-likelihood (used for softmax, greedy, top-k, and nucleus sampling), 
unlikelihood training \citep{welleck2020neural}, and entmax loss. For the unlikelihood training objective we replicated the authors' experiments. However, due to GPU memory constraints we had to reduce the context size from 512 to 256.  The hyperparameters were chosen based on a grid search over $\alpha \in \{1.1, 1.2, 1.3, 1.5\}$ for entmax sampling,  $k \in \{5, 10, 20, 50, 100\}$ for top-$k$ sampling, $P \in \{0.5, 0.8, 0.85, 0.9, 0.95, 0.97\}$ for nucleus sampling, and $\tau \in \{0.7,0.8,0.9,0.95,0.97\}$ for softmax with decreased temperature. The selected hyperparameters are reported in Table~\ref{hyperparam_lm}. We report the results obtained on the validation sets of WikiText-2, WikiText-103, and BookCorpus on Table~\ref{results_lm_auto_val}.
\begin{table}
\vspace{\baselineskip}
\centering \small
\setlength{\tabcolsep}{1ex}
\begin{tabular}{lccc}
\toprule
& WikiText-2 & WikiText-103 & BookCorpus\\
\midrule
$\alpha$ & $1.2$ & $1.2$ & $1.3$\\
$k$ & $50$ & $50$ & $20$\\
$P$ & $0.95$ & $0.95$ & $0.90$  \\
$\tau$ & $0.95$ & $0.95$ & $0.90$\\
\bottomrule
\end{tabular}
\caption{Values of hyperparameters selected for Language Modeling.}
\label{hyperparam_lm}
\end{table}
Additional settings and the computational infrastructure are described in App.~\ref{finetuning_lm}.

\paragraph{Results.}
Table~\ref{results_lm_auto} shows the results. We observe that entmax sampling achieves consistently better sparsemax scores and number of repetitions.  It also leads to better $\epsilon$-perplexity scores than all other methods except plain softmax, which attains similar scores (entmax is slightly better for 2 out of 3 datasets). 
The $\mathsf{JS}$ score appears to favor extremely sparse decoders, with greedy decoding achieving the best scores (but at the expense of many repetitions).\footnote{Figure~\ref{sweep} of App.~\ref{app:sweep} shows results on automatic metrics for top-$k$, nucleus, and entmax sampling on WikiText-103 validation set for various $K$, $P$, and $\alpha$.}

To help understand why entmax leads to better sparsemax scores and fewer repetitions,  Table~\ref{word_statistics} shows the mean, median, standard deviation, minimum, and maximum number of tokens each decoding strategy considers when predicting each word,  on the Wikitext-103 test set.
\begin{table}[t]
\vspace{\baselineskip}
\centering \small
\setlength{\tabcolsep}{1.2ex}
\resizebox{\columnwidth}{!}{
\begin{tabular}{l@{\hspace{2ex}}ccccc}
\toprule
     & \textsc{Mean} & \textsc{Median} & \textsc{SD} & \textsc{Min} & \textsc{Max}\\
\midrule
Softmax & 50,257 & 50,257 & 0 & 50,257 & 50,257 \\
Softmax-$\tau$ & 50,257 & 50,257 & 0 & 50,257 & 50,257 \\
Greedy &  1 & 1 & 0 & 1 & 1\\
Top-$k$  & 50 & 50 & 0 & 50 & 50\\
Nucleus & 562 & 210 & 1,187 & 1 & 19,945 \\
Entmax & 2,532 & 1,210 & 2,643 & 1 & 28,364 \\
\bottomrule
\end{tabular}}
\caption{Mean, median, standard deviation, minimum, and maximum number of tokens considered by each decoding method on the Wikitext-103 test set.} 
\label{word_statistics}
\end{table}
We  see that entmax sampling and nucleus sampling consider a lot more tokens than greedy decoding and top-$k$ sampling, which may be the reason for the smaller number of repetitions. 
A possible explanation for entmax sampling outperforming nucleus sampling is its higher standard deviation, suggesting that its sparsity range is more adaptive to the context. 

\paragraph{Ablation study.}
In order to understand whether the improved performance is caused by the mitigation of the sparsity mismatch between training and test times, we experimented (i) decoding with the entmax sampling method from a language model fine-tuned with negative log-likelihood,  and (ii) decoding with top-$k$ sampling and nucleus sampling from a model fine-tuned with the entmax loss. We conducted these experiments on the WikiText-103 dataset. 
\begin{table}
\vspace{\baselineskip}
\centering \small
\setlength{\tabcolsep}{1.6ex}
\begin{tabular}{l@{\hspace{3ex}}ccccc}
\toprule
     & \textsf{sp} & \textsf{JS} & $\epsilonppl$ & \textsc{rep} & \textsc{wrep}\\
\midrule
\multicolumn{6}{l}{Training with NLL} \\
\midrule
Top-$k$ & .683 & \textbf{.364} & 21.90  & .373 & .181 \\
Nucleus & .686 & .370 & 15.51 & .357 & .167 \\
Entmax &  .670 & .378 & 20.69 & .365 & .183\\
\midrule
\multicolumn{6}{l}{Training with Entmax Loss} \\
\midrule
Top-$k$ & .677 & .384 & 46.58 & .364 & .196  \\
Nucleus & .668 & .373 & 43.19 & .350 & .172 \\
Entmax & \textbf{.694} & .373 & \textbf{13.23} & \textbf{.346} & \textbf{.160} \\
\bottomrule
\end{tabular}
\caption{Language modeling ablation study on WikiText-103 test set.} 
\label{ablation}
\end{table}

As shown in Table~\ref{ablation}, our proposed approach, which decodes with entmax sampling from a model also fine-tuned with the entmax loss, is the one which leads to the best scores, as we see a considerable degradation when entmax is only used at training or at decoding time. This corroborates our hypothesis that the improved results come from eliminating the mismatch between training and decoding.

\subsection{Story completion} 
\label{sc_exp}
Next, we analyze the model's ability to generate long sequences of text using different sampling methods.\footnote{Softmax sampling is not considered since it has been shown to generate degenerate text \citep{holtzman2020curious}.} We performed completion of stories from the WritingPrompts dataset \citep{fan2018hierarchical}, using the models fine-tuned on  BookCorpus. WritingPrompts is a collection of human-written stories paired with writing prompts. We randomly selected 1,000 stories which were at least 200 words long and used the first 50 words as context for the models. Examples of stories generated with each method (Table~\ref{example} and Table~\ref{examples_stories} of App.~\ref{app:story_examples}) suggest that entmax sampling leads to more engaging stories while preventing degenerate text.
\begin{figure}[t]
\begin{center}
\includegraphics[width=\columnwidth]{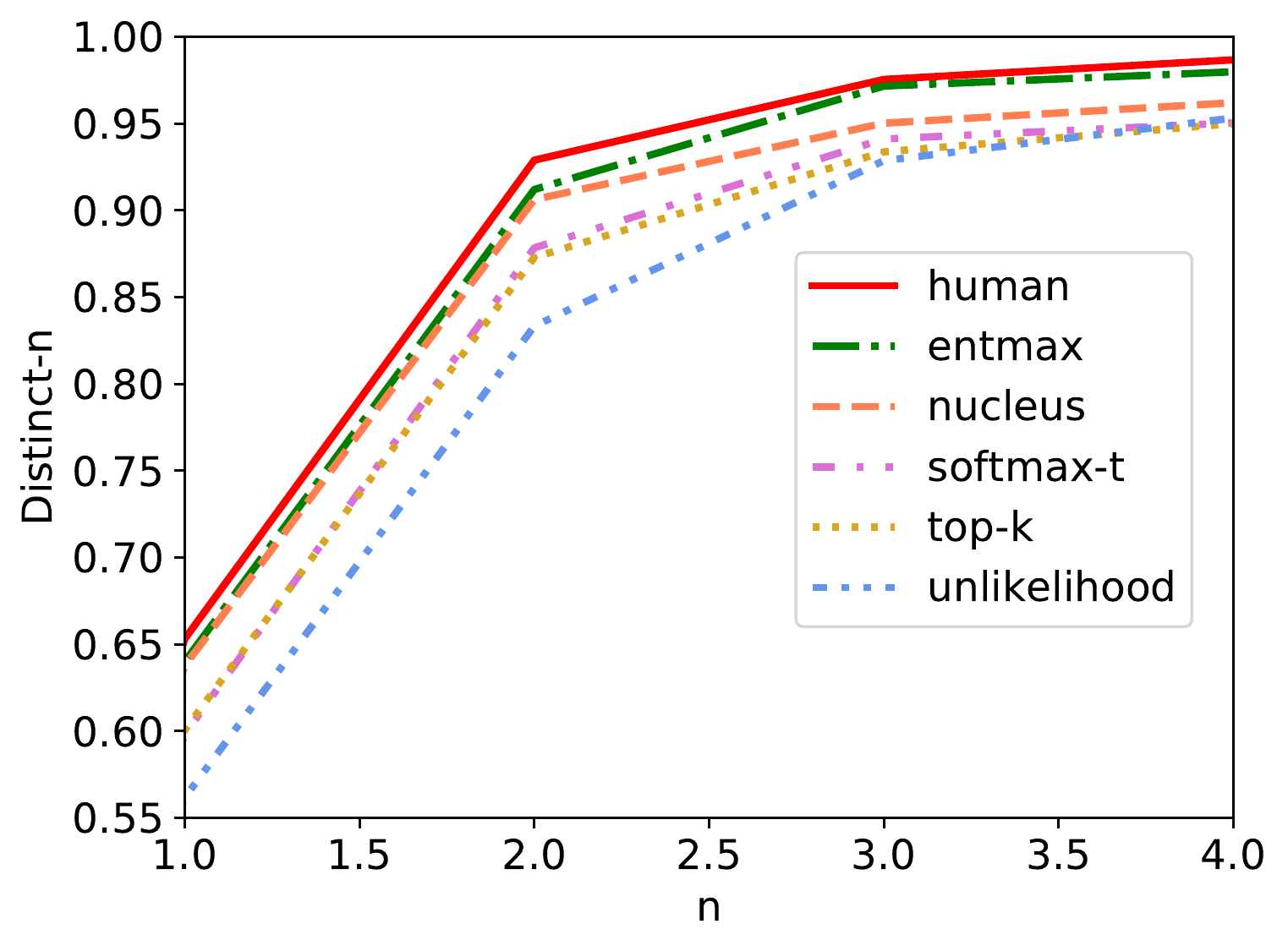}
\end{center}
\caption{Plot of the distinct-n metric for each sampling method on story completion, with $n=\{1,2,3,4\}$. The distinct-n results for greedy decoding are not shown since they are very small ($0.25$ for distinct-$4$).}
\label{distinct_lm_plot}
\end{figure}
To measure the stories' word diversity, we show in Figure~\ref{distinct_lm_plot} the distinct-$n$ metric\footnote{Distinct-$n$ corresponds to the number of distinct $n$-grams divided by the total number of generated words.} \citep{lidiversity} for the stories generated by each model. It can be seen that entmax sampling leads to more diverse unique n-grams for $n \in \{1, 2, 3, 4\}$, closer to human generated text.
We also measured the number of unique words in the stories generated: entmax sampling generated 14,702 different words, while softmax with decreased temperature, greedy decoding, top-$k$, nucleus sampling, and unlikelihood generated 12,447, 1,750, 11,803, 12,008, and 5,509 words, respectively. As expected, entmax leads to higher word diversity on par with human stories, which contain 15,377 different words.

\paragraph{Human evaluation.}
We performed human evaluation of greedy decoding, unlikelihood training, top-$k$, nucleus, and entmax sampling on completion of stories from the WritingPrompts datasets. We randomly selected 100 stories to perform the human evaluation. For each story, 5 judges from Amazon Mechanical Turk evaluated the story completions in 3 metrics: fluency (whether the text is syntactically and semantically correct), coherence (whether  the  story  continuation is related to the provided context and is consistent), and engagement (whether the annotator felt interested in the story). 
Ratings were given on a 5-point Likert scale, and the mean for each metric is reported in Table~\ref{results_lm_human}. 
Further details, including a screenshot of the annotator interface, are described in App.~\ref{human_evaluation_details_stories}.
\begin{table}[t]
\vspace{\baselineskip}
\centering \small
\setlength{\tabcolsep}{1ex}
\resizebox{\columnwidth}{!}{
\begin{tabular}{l@{\hspace{2.5ex}}cccc}
\toprule
     & \textsc{Fluency} & \textsc{Coherence} & \textsc{Engagement}\\
\midrule
Greedy & 2.5 & 2.3 & 2.3\\
top-$k$  & 3.3 & 2.9 & 2.9  \\
Nucleus  & \textbf{3.5} & 3.1 & 3.2 \\
Unlikelihood & 3.3 & 3.0 & 3.2 \\
Entmax & \textbf{3.5} & \textbf{3.2} & \textbf{3.6}\\
\bottomrule
\end{tabular}}
\caption{Human evaluation of story completion. All scores marked in bold at each column outperform the others with statistical significance, according to the Wilcoxon's test 
with p-value~$< 0.01$. The inter-annotator agreement (Fleiss Kappa) is $0.45$ for fluency, $0.41$ for coherence, and $0.63$ for engagement.}
\label{results_lm_human}
\end{table}
We observe that entmax sampling outperforms all other methods on coherence and engagement, having similar scores as nucleus sampling on fluency.

\subsection{Dialogue Generation}
\label{rp_exp}
To evaluate the sampling methods in an interactive setting, we experiment with dialogue generation. 
Its goal  
is to generate an utterance, given a context consisting of the previous utterances in the dialogue and, in some cases, initial context sentences with related information that can be describing personas, knowledge, or scenarios.

\paragraph{Datasets and metrics.}
We performed experiments with the PersonaChat dataset \citep{zhangpersonalizing}. It is a crowd-sourced dialogue dataset in which speakers were asked to condition their utterances on predefined personas. It contains 164,356 utterances over 10,981 dialogues. As there is no public test set, we report results on the validation set. 
We evaluate the word $F_1$-score, $\epsilon$-perplexity, sparsemax score, and Jensen-Shannon divergence. As for the language modeling experiments, $\epsilon$-perplexity, sparsemax score, and Jensen-Shannon are computed at the BPE level. 
We also report distinct-$n$ metric for $n=\{1,2\}$ 
and analyze how the models behave in dialogue simulations between two agents \citep{li2016deep}.

\paragraph{Fine-tuning GPT-2.}
In order to apply GPT-2 medium to the dialogue generation task, we follow \citet{wolf2019transfertransfo} and \citet{budzianowski2019hello}: the input given to the language model consists of the sentences describing the persona the model should impersonate, and the history utterances. In order for the model to adapt to dialogue, the word and position embeddings are augmented with dialogue-state embeddings that indicate whether tokens are from a persona sentence, speaker 1, or speaker 2. These embeddings are learned during fine-tuning.
The hyperparameters $\alpha$, $k$, $P$, and $\tau$ were chosen with a grid search over the sets of values $\alpha \in \{1.2, 1.3, 1.5, 2\}$, $k\in\{5, 10, 20, 50, 100\}$, $P\in\{0.5, 0.8, 0.85, 0.9, 0.95\}$, and $\tau \in \{0.7,0.8,0.9,0.95\}$, using the sparsemax score. The values chosen are $1.5$, $10$, $0.9$, and $0.8$, respectively.
Additional settings are described in App.~\ref{finetuning_dialogue}.

\paragraph{Automatic metrics results.}
We report the results in Table~\ref{results_dialog_auto}. Entmax again outperforms all the other methods in sparsemax score and $\epsilon$-perplexity. It also has the lowest $\mathsf{JS}$ (same as top-$k$ and softmax-$\tau$). Entmax also leads to fewer repetitions, having higher distinct-1 and distinct-2 scores. However, its F$_1$ score is lower (similar findings have been reported in \citet{li2019don}). This can be due to dialogue generation being an open-ended generation task that can have multiple correct answers. 

\begin{table}[t]
\vspace{\baselineskip}
\centering \small
\setlength{\tabcolsep}{0.9ex}
\resizebox{\columnwidth}{!}{
\begin{tabular}{l@{\hspace{1ex}}cccccc}
\toprule
    & \textsf{sp} & \textsc{JS} & $\epsilonppl$ & $\mathsf{F}_1$ & \textsc{dist-1} & \textsc{dist-2} \\
\midrule
Softmax & 0.636 & 0.412 & 17.21 & 14.21 & 0.4325 & 0.8422  \\
Softmax-$\tau$ & 0.621 & \textbf{0.393} & 17.18 & 16.31 & 0.4312 & 0.8289 \\
Greedy & 0.422 & 0.401 & 1031.79 & \textbf{21.79} & 0.4305 & 0.7958\\
Top-$k$  & 0.626 & \textbf{0.393} &  47.79 & 17.34	& 0.4378 &	0.8344 \\
Nucleus & 0.632 & 0.399  & 26.98 & 15.98 & 0.4334 &	0.8436 \\
Entmax & \textbf{0.642} & \textbf{0.393} &  \textbf{17.10} & 15.02 & \textbf{0.4532} & \textbf{0.8494} \\
\bottomrule
\end{tabular}}
\caption{Automatic evaluation of dialogue generation on the Persona-Chat validation set.}
\label{results_dialog_auto}
\end{table}

Additionally, we simulated a conversation between two agents of the same model~\citep{li2016deep}. We chose different personas randomly for the two agents. Then a first utterance from the PersonaChat dataset was given as context. Some conversation examples are presented in Tables \ref{examples_dialogue_1}-\ref{examples_dialogue_4} of App.~\ref{app:sim_examples}. We measured the average length of conversations, considering that the conversation is finished when utterances overlap 80$\%$ or more, when there is no response by an agent, or when it reaches 20 utterances (similar procedure as \citet{li2016deep}). 
We also measured the number of unique words, and the distinct-$n$ metric for $n=\{1,2\}$.
\begin{table}[t]
\vspace{\baselineskip}
\centering \small
\setlength{\tabcolsep}{1ex}
\resizebox{\columnwidth}{!}{
\begin{tabular}{l@{\hspace{1.3ex}}ccccc}
\toprule
     & \textsc{Length} & \textsc{Unique words} &  \textsc{dist-1} & \textsc{dist-2} \\
\midrule
Softmax  & 13.98 & 11,242 & 0.6084 & 0.8824 \\
Softmax-$\tau$ &  14.82 &  5,495 &  0.5384 & 0.6936\\
Greedy & 7.83 & 4,229 & 0.4853 & 0.6732	 \\
Top-$k$  & 14.72 & 8,833 & 0.5623 & 0.8461 \\
Nucleus & 15.56 & 10,098  & 0.5836 & 0.8728 \\
Entmax &  \textbf{15.83} & \textbf{13,020} & \textbf{0.6546} & \textbf{0.9211} \\
\bottomrule
\end{tabular}}
\caption{Evaluation of dialogue simulations between two agents using the different sampling methods.}
\label{results_dialog_model_model}
\end{table}
As shown in Table~\ref{results_dialog_model_model}, entmax sampling leads to longer conversations with higher word diversity and higher number of distinct 1-grams and 2-grams.

\paragraph{Human evaluation.}
Finally, we performed human evaluation following the ConvAI2 challenge:  12 volunteers had 30 conversations each with models using the different sampling methods. The volunteers scored the conversations from 1 to 5 in terms of fluency, consistency (whether the model’s utterances are coherent with their persona and the model does not contradict itself), and engagement. The model’s personas  were  randomly  selected  from  the  PersonaChat  validation  set.
Results are reported in Table~\ref{results_dialog_human}. 
\begin{table}[t]
\vspace{\baselineskip}
\centering \small
\setlength{\tabcolsep}{1ex}
\begin{tabular}{l@{\hspace{2.5ex}}cccc}
\toprule
     & \textsc{Fluency} & \textsc{Consistency} & \textsc{Engagement}\\
\midrule
Greedy & \textbf{4.1} & 3.0 & 2.5 \\ 
Top-$k$ & \textbf{4.0} & 3.2 & 3.3\\
Nucleus  & \textbf{4.1} & 3.4 & 3.3  \\
Entmax & \textbf{4.1} & \textbf{3.6} & \textbf{3.9}  \\
\bottomrule
\end{tabular}
\caption{Human evaluation of dialogue generation. All scores marked in bold at each column outperform the non-bold ones with statistical significance, according to the Wilcoxon's test with p-value~$< 0.01$. 
\label{results_dialog_human}}
\end{table}
Entmax sampling outperforms the other methods in consistency and engagement, having similar scores in fluency. This means entmax sampling does not only generate the most interesting conversation utterances, but it also leads to an improvement of the conversation consistency.

\section{Conclusions}
We proposed entmax sampling as a new strategy for generating text from a sparse probability distribution. 
It provides three main advantages: (i) it offers a natural way of sampling directly from the output probability distribution; 
 (ii) the distribution sparsity is modeled during training, avoiding a sparsity mismatch between training and run time; 
(iii) when sampling with entmax, the number of words to be considered varies with the context, as in nucleus sampling and in contrast to top-$k$ sampling. 
Additionally, we proposed new metrics for evaluating language models that produce sparse and truncated probability distributions: $\epsilon$-perplexity, sparsemax score, and Jensen-Shannon divergence.

Experiments show that entmax sampling
leads to higher n-gram diversity, fewer repetitions, and similar or improved results in automatic metrics.
Human evaluation confirms that entmax outperforms greedy decoding, top-$k$, and nucleus sampling in coherence/consistency and engagement, and is similar or better in terms of fluency.

\section*{Acknowledgements}
This work was supported by the European Research Council (ERC StG DeepSPIN 758969), 
by the P2020 project MAIA (contract 045909), 
 and by the Funda\c{c}\~ao para a Ci\^encia e Tecnologia 
through contract UIDB/50008/2020 and contract PD/BD/150633/2020 in the scope of the  Doctoral Program  FCT - PD/00140/2013 NETSyS, ``Networked Interactive Cyber Physical System''. We thank Ben Peters, Erick Fonseca, Gon\c{c}alo Correia, Marcos Treviso, Tsvetomila Mihaylova, Vlad Niculae, and the reviewers for helpful discussion and feedback.

\bibliography{anthology,emnlp2020}

\begin{thebibliography}{45}
\expandafter\ifx\csname natexlab\endcsname\relax\def\natexlab#1{#1}\fi

\bibitem[{Bahdanau et~al.(2015)Bahdanau, Cho, and Bengio}]{Bahdanau2015}
Dzmitry Bahdanau, Kyunghyun Cho, and Yoshua Bengio. 2015.
\newblock \href{https://arxiv.org/pdf/1409.0473.pdf}{Neural Machine Translation
  by Jointly Learning to Align and Translate}.
\newblock In \emph{Proc. ICLR}.

\bibitem[{Banerjee et~al.(2005)Banerjee, Merugu, Dhillon, and
  Ghosh}]{banerjee2005clustering}
Arindam Banerjee, Srujana Merugu, Inderjit~S Dhillon, and Joydeep Ghosh. 2005.
\newblock
  \href{http://www.jmlr.org/papers/volume6/banerjee05b/banerjee05b.pdf}{Clustering
  with Bregman divergences}.
\newblock \emph{Journal of machine learning research}.

\bibitem[{Blondel et~al.(2019)Blondel, Martins, and
  Niculae}]{blondel2019learning}
Mathieu Blondel, Andre Martins, and Vlad Niculae. 2019.
\newblock \href{https://arxiv.org/pdf/1805.09717.pdf}{Learning Classifiers with
  Fenchel-Young Losses: Generalized Entropies, Margins, and Algorithms}.
\newblock In \emph{Proc. AISTATS}.

\bibitem[{Budzianowski and Vuli{\'c}(2019)}]{budzianowski2019hello}
Pawe{\l} Budzianowski and Ivan Vuli{\'c}. 2019.
\newblock \href{https://www.aclweb.org/anthology/D19-5602.pdf}{Hello, It’s
  GPT-2-How Can I Help You? Towards the Use of Pretrained Language Models for
  Task-Oriented Dialogue Systems}.
\newblock In \emph{Proc. Workshop on Neural Generation and Translation}.

\bibitem[{Chen and Goodman(1999)}]{chen1999empirical}
Stanley~F Chen and Joshua Goodman. 1999.
\newblock \href{https://www.aclweb.org/anthology/P96-1041.pdf}{An empirical
  study of smoothing techniques for language modeling}.
\newblock \emph{Computer Speech \& Language}.

\bibitem[{Chorowski et~al.(2015)Chorowski, Bahdanau, Serdyuk, Cho, and
  Bengio}]{chorowski2015attention}
Jan~K Chorowski, Dzmitry Bahdanau, Dmitriy Serdyuk, Kyunghyun Cho, and Yoshua
  Bengio. 2015.
\newblock
  \href{https://papers.nips.cc/paper/5847-attention-based-models-for-speech-recognition.pdf}{Attention-based
  models for speech recognition}.
\newblock In \emph{Proc. NIPS}.

\bibitem[{C{\'\i}fka et~al.(2018)C{\'\i}fka, Severyn, Alfonseca, and
  Filippova}]{cifka2018eval}
Ond{\v{r}}ej C{\'\i}fka, Aliaksei Severyn, Enrique Alfonseca, and Katja
  Filippova. 2018.
\newblock \href{https://arxiv.org/pdf/1804.07972.pdf}{Eval all, trust a few, do
  wrong to none: Comparing sentence generation models}.
\newblock \emph{arXiv preprint arXiv:1804.07972}.

\bibitem[{Dai et~al.(2019)Dai, Yang, Yang, Cohen, Carbonell, Le, and
  Salakhutdinov}]{dai2019transformer}
Zihang Dai, Zhilin Yang, Yiming Yang, William~W Cohen, Jaime Carbonell, Quoc~V
  Le, and Ruslan Salakhutdinov. 2019.
\newblock \href{https://www.aclweb.org/anthology/P19-1285.pdf}{Transformer-xl:
  Attentive language models beyond a fixed-length context}.
\newblock \emph{Proc. ACL}.

\bibitem[{Deza and Deza(2009)}]{deza2009encyclopedia}
Michel~Marie Deza and Elena Deza. 2009.
\newblock Encyclopedia of distances.
\newblock In \emph{Encyclopedia of distances}, pages 1--583. Springer.

\bibitem[{Fan et~al.(2018)Fan, Lewis, and Dauphin}]{fan2018hierarchical}
Angela Fan, Mike Lewis, and Yann Dauphin. 2018.
\newblock \href{https://www.aclweb.org/anthology/P18-1082.pdf}{Hierarchical
  Neural Story Generation}.
\newblock In \emph{Proc. ACL}.

\bibitem[{Ficler and Goldberg(2017)}]{ficler2017controlling}
Jessica Ficler and Yoav Goldberg. 2017.
\newblock \href{https://www.aclweb.org/anthology/W17-4912.pdf}{Controlling
  Linguistic Style Aspects in Neural Language Generation}.
\newblock In \emph{Proc. of the Workshop on Stylistic Variation}.

\bibitem[{Grosse et~al.(2002)Grosse, Bernaola-Galv{\'a}n, Carpena,
  Rom{\'a}n-Rold{\'a}n, Oliver, and Stanley}]{grosse2002analysis}
Ivo Grosse, Pedro Bernaola-Galv{\'a}n, Pedro Carpena, Ram{\'o}n
  Rom{\'a}n-Rold{\'a}n, Jose Oliver, and H~Eugene Stanley. 2002.
\newblock
  \href{https://pdfs.semanticscholar.org/31e0/71bac34c36ecd8bba919c4f3a8d5514fe710.pdf}{Analysis
  of symbolic sequences using the Jensen-Shannon divergence}.
\newblock \emph{Physical Review E}.

\bibitem[{Holtzman et~al.(2018)Holtzman, Buys, Forbes, Bosselut, Golub, and
  Choi}]{holtzman2018learning}
Ari Holtzman, Jan Buys, Maxwell Forbes, Antoine Bosselut, David Golub, and
  Yejin Choi. 2018.
\newblock \href{https://www.aclweb.org/anthology/P18-1152.pdf}{Learning to
  Write with Cooperative Discriminators}.
\newblock In \emph{Proc. ACL}.

\bibitem[{Holtzman et~al.(2020)Holtzman, Buys, Forbes, and
  Choi}]{holtzman2020curious}
Ari Holtzman, Jan Buys, Maxwell Forbes, and Yejin Choi. 2020.
\newblock \href{https://openreview.net/pdf?id=rygGQyrFvH}{The curious case of
  neural text degeneration}.
\newblock In \emph{Proc. ICLR}.

\bibitem[{Jelinek et~al.(1977)Jelinek, Mercer, Bahl, and
  Baker}]{jelinek1977perplexity}
Fred Jelinek, Robert~L Mercer, Lalit~R Bahl, and James~K Baker. 1977.
\newblock Perplexity-a measure of the difficulty of speech recognition tasks.
\newblock \emph{The Journal of the Acoustical Society of America}.

\bibitem[{Kingma and Ba(2015)}]{kingma2014adam}
Diederik~P Kingma and Jimmy Ba. 2015.
\newblock \href{https://arxiv.org/pdf/1412.6980.pdf}{Adam: A method for
  stochastic optimization}.
\newblock In \emph{Proc. ICLR}.

\bibitem[{Knowles et~al.(2016)Knowles, Renduchintala, Koehn, and
  Eisner}]{knowles2016analyzing}
Rebecca Knowles, Adithya Renduchintala, Philipp Koehn, and Jason Eisner. 2016.
\newblock \href{https://www.aclweb.org/anthology/K16-1013.pdf}{Analyzing
  Learner Understanding of Novel {L}2 Vocabulary}.
\newblock In \emph{Proc. SIGNLL}.

\bibitem[{Kulikov et~al.(2018)Kulikov, Miller, Cho, and
  Weston}]{kulikov2018importance}
Ilya Kulikov, Alexander~H Miller, Kyunghyun Cho, and Jason Weston. 2018.
\newblock \href{https://arxiv.org/pdf/1811.00907.pdf}{Importance of a search
  strategy in neural dialogue modelling}.
\newblock \emph{arXiv preprint arXiv:1811.00907}.

\bibitem[{Li et~al.(2016{\natexlab{a}})Li, Galley, Brockett, Gao, and
  Dolan}]{lidiversity}
Jiwei Li, Michel Galley, Chris Brockett, Jianfeng Gao, and Bill Dolan.
  2016{\natexlab{a}}.
\newblock \href{https://www.aclweb.org/anthology/N16-1014.pdf}{A
  Diversity-Promoting Objective Function for Neural Conversation Models}.
\newblock In \emph{Proc. NAACL}.

\bibitem[{Li et~al.(2016{\natexlab{b}})Li, Monroe, and Jurafsky}]{li2016simple}
Jiwei Li, Will Monroe, and Dan Jurafsky. 2016{\natexlab{b}}.
\newblock \href{https://arxiv.org/pdf/1611.08562.pdf}{A simple, fast diverse
  decoding algorithm for neural generation}.
\newblock \emph{arXiv preprint arXiv:1611.08562}.

\bibitem[{Li et~al.(2016{\natexlab{c}})Li, Monroe, Ritter, Jurafsky, Galley,
  and Gao}]{li2016deep}
Jiwei Li, Will Monroe, Alan Ritter, Dan Jurafsky, Michel Galley, and Jianfeng
  Gao. 2016{\natexlab{c}}.
\newblock \href{https://www.aclweb.org/anthology/D16-1127.pdf}{Deep
  Reinforcement Learning for Dialogue Generation}.
\newblock In \emph{Proc. EMNLP}.

\bibitem[{Li et~al.(2017)Li, Monroe, Shi, Jean, Ritter, and
  Jurafsky}]{li2017-adversarial}
Jiwei Li, Will Monroe, Tianlin Shi, S{\'e}bastien Jean, Alan Ritter, and Dan
  Jurafsky. 2017.
\newblock \href{https://www.aclweb.org/anthology/D17-1230.pdf}{Adversarial
  Learning for Neural Dialogue Generation}.
\newblock In \emph{Proc. EMNLP}.

\bibitem[{Li et~al.(2020)Li, Roller, Kulikov, Welleck, Boureau, Cho, and
  Weston}]{li2019don}
Margaret Li, Stephen Roller, Ilia Kulikov, Sean Welleck, Y-Lan Boureau,
  Kyunghyun Cho, and Jason Weston. 2020.
\newblock \href{https://arxiv.org/pdf/1911.03860.pdf}{Don't Say That! Making
  Inconsistent Dialogue Unlikely with Unlikelihood Training}.
\newblock In \emph{Proc. ACL}.

\bibitem[{Li et~al.(2019)Li, Weston, and Roller}]{li2019acute}
Margaret Li, Jason Weston, and Stephen Roller. 2019.
\newblock \href{https://arxiv.org/pdf/1909.03087.pdf}{Acute-eval: Improved
  dialogue evaluation with optimized questions and multi-turn comparisons}.
\newblock \emph{arXiv preprint arXiv:1909.03087}.

\bibitem[{Martins and Astudillo(2016)}]{martins2016softmax}
Andre Martins and Ramon Astudillo. 2016.
\newblock \href{http://proceedings.mlr.press/v48/martins16.pdf}{From softmax to
  sparsemax: A sparse model of attention and multi-label classification}.
\newblock In \emph{Proc. ICML}.

\bibitem[{Merity et~al.(2016)Merity, Xiong, Bradbury, and
  Socher}]{merity2016pointer}
Stephen Merity, Caiming Xiong, James Bradbury, and Richard Socher. 2016.
\newblock \href{https://arxiv.org/pdf/1609.07843.pdf}{Pointer Sentinel Mixture
  Models}.
\newblock In \emph{Proc. ICLR}.

\bibitem[{Papineni et~al.(2002)Papineni, Roukos, Ward, and
  Zhu}]{papineni2002bleu}
Kishore Papineni, Salim Roukos, Todd Ward, and Wei-Jing Zhu. 2002.
\newblock \href{https://www.aclweb.org/anthology/P02-1040.pdf}{BLEU: a method
  for automatic evaluation of machine translation}.
\newblock In \emph{Proc. ACL}.

\bibitem[{Patrick and Fischer(1969)}]{patrick1969nonparametric}
E~Patrick and F~Fischer. 1969.
\newblock Nonparametric feature selection.
\newblock \emph{IEEE Transactions on Information Theory}.

\bibitem[{Peters et~al.(2019)Peters, Niculae, and Martins}]{peters2019sparse}
Ben Peters, Vlad Niculae, and Andr{\'e}~FT Martins. 2019.
\newblock \href{https://www.aclweb.org/anthology/P19-1146.pdf}{Sparse
  Sequence-to-Sequence Models}.
\newblock In \emph{Proc. ACL}.

\bibitem[{Press et~al.(2017)Press, Bar, Bogin, Berant, and
  Wolf}]{press2017language}
Ofir Press, Amir Bar, Ben Bogin, Jonathan Berant, and Lior Wolf. 2017.
\newblock \href{https://arxiv.org/pdf/1706.01399.pdf}{Language generation with
  recurrent generative adversarial networks without pre-training}.
\newblock \emph{arXiv preprint arXiv:1706.01399}.

\bibitem[{Radford et~al.(2019)Radford, Wu, Child, Luan, Amodei, and
  Sutskever}]{radford2019language}
Alec Radford, Jeffrey Wu, Rewon Child, David Luan, Dario Amodei, and Ilya
  Sutskever. 2019.
\newblock
  \href{https://d4mucfpksywv.cloudfront.net/better-language-models/language_models_are_unsupervised_multitask_learners.pdf}{Language
  models are unsupervised multitask learners}.
\newblock \emph{OpenAI Blog}.

\bibitem[{Rush et~al.(2015)Rush, Chopra, and Weston}]{rush2015neural}
Alexander~M Rush, Sumit Chopra, and Jason Weston. 2015.
\newblock \href{https://www.aclweb.org/anthology/D15-1044.pdf}{A Neural
  Attention Model for Abstractive Sentence Summarization}.
\newblock In \emph{Proc. EMNLP}.

\bibitem[{Sennrich et~al.(2016)Sennrich, Haddow, and
  Birch}]{sennrich2016neural}
Rico Sennrich, Barry Haddow, and Alexandra Birch. 2016.
\newblock \href{https://www.aclweb.org/anthology/P16-1162.pdf}{Neural Machine
  Translation of Rare Words with Subword Units}.
\newblock In \emph{Proc.ACL}.

\bibitem[{Stahlberg and Byrne(2019)}]{stahlberg2019nmt}
Felix Stahlberg and Bill Byrne. 2019.
\newblock \href{https://www.aclweb.org/anthology/D19-1331.pdf}{On NMT Search
  Errors and Model Errors: Cat Got Your Tongue?}
\newblock In \emph{Proc. EMNLP}.

\bibitem[{Sutskever et~al.(2014)Sutskever, Vinyals, and
  Le}]{sutskever2014sequence}
I~Sutskever, O~Vinyals, and QV~Le. 2014.
\newblock
  \href{https://papers.nips.cc/paper/5346-sequence-to-sequence-learning-with-neural-networks.pdf}{Sequence
  to sequence learning with neural networks}.
\newblock \emph{Proc. NIPS}.

\bibitem[{Tsallis(1988)}]{tsallis1988possible}
Constantino Tsallis. 1988.
\newblock Possible generalization of boltzmann-gibbs statistics.
\newblock \emph{Journal of statistical physics}.

\bibitem[{Vaswani et~al.(2017)Vaswani, Shazeer, Parmar, Uszkoreit, Jones,
  Gomez, Kaiser, and Polosukhin}]{vaswani2017attention}
Ashish Vaswani, Noam Shazeer, Niki Parmar, Jakob Uszkoreit, Llion Jones,
  Aidan~N Gomez, {\L}ukasz Kaiser, and Illia Polosukhin. 2017.
\newblock
  \href{https://papers.nips.cc/paper/7181-attention-is-all-you-need.pdf}{Attention
  is all you need}.
\newblock In \emph{Proc. NIPS}.

\bibitem[{Vijayakumar et~al.(2018)Vijayakumar, Cogswell, Selvaraju, Sun, Lee,
  Crandall, and Batra}]{vijayakumar2018diverse}
Ashwin~K Vijayakumar, Michael Cogswell, Ramprasaath~R Selvaraju, Qing Sun,
  Stefan Lee, David Crandall, and Dhruv Batra. 2018.
\newblock
  \href{http://web.engr.oregonstate.edu/~leestef/pdfs/diversebeam2018aaai.pdf}{Diverse
  beam search for improved description of complex scenes}.
\newblock In \emph{Proc. AAAI}.

\bibitem[{Welleck et~al.(2020)Welleck, Kulikov, Roller, Dinan, Cho, and
  Weston}]{welleck2020neural}
Sean Welleck, Ilia Kulikov, Stephen Roller, Emily Dinan, Kyunghyun Cho, and
  Jason Weston. 2020.
\newblock \href{https://openreview.net/pdf?id=SJeYe0NtvH}{Neural text
  generation with unlikelihood training}.
\newblock In \emph{Proc. ICLR}.

\bibitem[{Wolf et~al.(2019)Wolf, Sanh, Chaumond, and
  Delangue}]{wolf2019transfertransfo}
Thomas Wolf, Victor Sanh, Julien Chaumond, and Clement Delangue. 2019.
\newblock \href{https://arxiv.org/pdf/1901.08149.pdf}{Transfertransfo: A
  transfer learning approach for neural network based conversational agents}.
\newblock \emph{arXiv preprint arXiv:1901.08149}.

\bibitem[{Xu et~al.(2018)Xu, Ren, Lin, and Sun}]{xu2018diversity}
Jingjing Xu, Xuancheng Ren, Junyang Lin, and Xu~Sun. 2018.
\newblock
  \href{https://www.aclweb.org/anthology/D18-1428.pdf}{Diversity-promoting gan:
  A cross-entropy based generative adversarial network for diversified text
  generation}.
\newblock In \emph{Proc. EMNLP}.

\bibitem[{Yu et~al.(2017)Yu, Zhang, Wang, and Yu}]{yu2017seqgan}
Lantao Yu, Weinan Zhang, Jun Wang, and Yong Yu. 2017.
\newblock \href{https://arxiv.org/pdf/1609.05473.pdf}{Seqgan: Sequence
  generative adversarial nets with policy gradient}.
\newblock In \emph{Proc. AAAI}.

\bibitem[{Zhang et~al.(2018)Zhang, Dinan, Urbanek, Szlam, Kiela, and
  Weston}]{zhangpersonalizing}
Saizheng Zhang, Emily Dinan, Jack Urbanek, Arthur Szlam, Douwe Kiela, and Jason
  Weston. 2018.
\newblock \href{https://www.aclweb.org/anthology/P18-1205.pdf}{Personalizing
  Dialogue Agents: {I} have a dog, do you have pets too?}
\newblock In \emph{Proc. ACL)}.

\bibitem[{Zhu et~al.(2018)Zhu, Lu, Zheng, Guo, Zhang, Wang, and
  Yu}]{zhu2018texygen}
Yaoming Zhu, Sidi Lu, Lei Zheng, Jiaxian Guo, Weinan Zhang, Jun Wang, and Yong
  Yu. 2018.
\newblock \href{https://arxiv.org/pdf/1802.01886.pdf}{Texygen: A benchmarking
  platform for text generation models}.
\newblock In \emph{Proc. SIGIR}.

\bibitem[{Zhu et~al.(2015)Zhu, Kiros, Zemel, Salakhutdinov, Urtasun, Torralba,
  and Fidler}]{Zhu2015AligningBA}
Yukun Zhu, Ryan Kiros, Richard~S. Zemel, Ruslan Salakhutdinov, Raquel Urtasun,
  Antonio Torralba, and Sanja Fidler. 2015.
\newblock
  \href{https://www.cv-foundation.org/openaccess/content_iccv_2015/papers/Zhu_Aligning_Books_and_ICCV_2015_paper.pdf}{Aligning
  Books and Movies: Towards Story-Like Visual Explanations by Watching Movies
  and Reading Books}.
\newblock In \emph{Proc. ICCV}.

\end{thebibliography}
\bibliographystyle{acl_natbib}

\clearpage
\appendix
\begin{center}
{\LARGE \textbf{Supplementary Material}}
\end{center}

\section{Selection of the optimal $\epsilon$ for the $\epsilon$-perplexity}
\label{epsilon_opt}
We show here that the optimal $\epsilon$ for the computation of the $\epsilon$-perplexity for each decoding method can be easily obtained by solving a convex optimization problem.

For a given $\epsilon$, which can be reparametrized as $\lambda = \frac{\epsilon |\mathcal{V}|}{1 + \epsilon |\mathcal{V}|} = \frac{1}{1 + (\epsilon |\mathcal{V}|)^{-1}} \in [0,1]$, the average negative log-likelihood on a validation set is:
\begin{align}
    F(\lambda) &= -\frac{1}{T}\sum_{t=1}^T \log \left( (1-\lambda)p_{\theta}(x_t) + \frac{\lambda}{|\mathcal{V}|} \right) \nonumber\\
    &= -\frac{1}{T}\sum_{t=1}^T \log (a_t \lambda + b_t),
\end{align}
where $a_t = |\mathcal{V}|^{-1} - p_{\theta}(x_t)$ and $b_t = p_{\theta}(x_t)$. 
The function $F$ is the composition of a convex function with an affine function, hence it is convex. Therefore it has a  global minimum. 
Its derivative is:
\begin{equation}
F'(\lambda) = -\frac{1}{T}\sum_{t=1}^T \frac{a_t}{a_t \lambda + b_t}.
\end{equation}
Since we constrain $\lambda \in [0,1]$, we can obtain the optimal $\lambda$ by initializing with $\lambda=0.5$ and iterating the following projected gradient rule:
\begin{equation}
    \lambda \leftarrow \max\{0, \min\{1, \lambda - \eta F'(\lambda) \} \}.
\end{equation}
where $\eta$ is a stepsize. 
Since $\lambda =  \frac{1}{1 + (\epsilon |\mathcal{V}|)^{-1}}$, we can invert this equation to obtain the optimal $\epsilon$  as $\epsilon = \frac{1}{|\mathcal{V}|(\lambda^{-1} - 1)} = \frac{\lambda}{|\mathcal{V}|(1 - \lambda)}$.

\section{Proof of boundedness of the sparsemax score}\label{sec:proof_sparsemax_bounded}
We show here that the sparsemax score in Eq.~\ref{eq:sparsemax_score} is always bounded between 0 and 1.

The fact that $\mathsf{sp} \le 1$ simply follows from the fact \citep[Prop.~2]{blondel2019learning} that 
any Fenchel-Young loss (which includes $\ell_2(\bs{z}, x)$) is non-negative. Since  $\mathsf{sp} = 1 - \min \{\ell_2(\bs{z}, x) \mid \mathsf{sparsemax}(\bs{z}) = \bs{p}_\theta\}$, it follows that $\mathsf{sp}\le 1$. 
Let us see when the maximal value 1 is attained. 
We have:
\begin{align}\label{eq:sparsemax_bound}
    \mathsf{sp} &= {p}_\theta(x) + \mathsf{H}_2(\bs{p}_\theta)\nonumber\\
    &= {p}_\theta(x) + \frac{1}{2}(1 - \|\bs{p}_\theta\|^2)\nonumber\\
    &= -\frac{1}{2}{p}_\theta(x)^2 + {p}_\theta(x) - \frac{1}{2}\sum_{x'\ne x} {p}_\theta(x')^2 + \frac{1}{2}\nonumber\\
    &= -\frac{1}{2}({p}_\theta(x) - 1)^2 - \frac{1}{2}\sum_{x'\ne x} {p}_\theta(x')^2 +1.
\end{align}
Since the Gini entropy is maximized by the uniform distribution, the maximum distribution in Eq.~\ref{eq:sparsemax_bound} is of the form $\bs{p}_\theta = \left(1-t, \frac{t}{|\mathcal{V}|-1}, \ldots, \frac{t}{|\mathcal{V}|-1}\right)$ for $t \in [0,1]$. 
Replacing in Eq.~\ref{eq:sparsemax_bound}, we obtain
\begin{align}
    \mathsf{sp} &= -\frac{1}{2}t^2 - \frac{1}{2}\frac{t^2}{|\mathcal{V}|-1} +1\nonumber\\
    &= 1- \frac{t^2}{2}\left(1+\frac{1}{|\mathcal{V}|-1}\right).
\end{align}
This is maximized by $t=0$, which corresponds to $\bs{p}_\theta = \bs{e}_x$.

To see that we always have $\mathsf{sp} \ge 0$, we use the fact that the Gini entropy $\mathsf{H}_2(\bs{p}_\theta)$ is always non-negative (zero if and only if $\bs{p}_\theta$ is a one-hot distribution), which is clear from the definition in footnote~\ref{foot:gini}, and that $p(x) \ge 0$; therefore, the sum of these two terms is also non-negative, and zero if and only if $\bs{p}_\theta = \bs{e}_x'$ with $x' \ne x$. 

\section{Relation between Patrick-Fischer distance and sparsemax score}
\label{sec:patrick_fischer}
We show here that the sparsemax score is equivalent to one minus the one half of the squared Patrick-Fisher distance between the distribution probability over the words $\bs{p}_\theta$ and the indicator one-hot vector $\bs{e}_x$ which corresponds to the ground truth word $x$. 

The Patrick-Fischer distance between two distributions is $D_{\mathrm{PF}}(\bs{p}, \bs{q}) = \|\bs{p} - \bs{q}\|_2$. 
We have:
\begin{eqnarray}
    &&1 - \frac{1}{2} D^2_{\mathrm{PF}}(\bs{p}_\theta, \bs{e}_x) \nonumber\\
    &=& 1 - \frac{1}{2}\sum_{x'} (p_\theta(x') - e_{x}(x'))^2\nonumber\\
    &=& 1 - \frac{1}{2}\sum_{x'\ne x} p_\theta(x')^2 - \frac{1}{2}(1 - p_\theta(x))^2 \nonumber\\
    &=& 1 - \frac{1}{2}\sum_{x'} p_\theta(x')^2 + \frac{1}{2}p_\theta(x)^2 - \frac{1}{2} \nonumber\\ &&- \frac{1}{2}p_\theta(x)^2 + p_\theta(x) \nonumber\\
    &=&  \frac{1}{2} - \frac{1}{2}\sum_{x'} p_\theta(x')^2 + p_\theta(x) \nonumber\\
    &=& p_\theta(x) + \mathsf{H}_2(\bs{p}_\theta),
\end{eqnarray}
which equals the sparsemax score defined in Eq.~\ref{eq:sparsemax_score}.

\section{$\mathsf{JS}$ divergence as a language model metric}
\label{js_}

The Jensen-Shannon divergence between the model  probability distribution over the words $\bs{p}_\theta$ and the indicator one-hot vector $\bs{e}_x$ which corresponds to the ground truth word $x$ can be defined as:
\begin{eqnarray}
    &&\mathsf{JS}(\bs{p}_\theta, \bs{e}_x) \nonumber\\
    &=& \frac{1}{2}\mathsf{KL}\left(\bs{p}_\theta \| \frac{\bs{p}_\theta + \bs{e}_x}{2}\right) \nonumber\\
    &=& \mathsf{H}\left(\frac{\bs{p}_\theta + \bs{e}_x}{2}\right) - \frac{1}{2}\mathsf{H}(\bs{p}_\theta) - \underbrace{\frac{1}{2}\mathsf{H}(\bs{e}_x)}_{= 0} \nonumber\\
    &=& -\sum_{x' \ne x} \frac{p_\theta(x')}{2} \log \frac{p_\theta(x')}{2} \nonumber\\&&- \frac{1+p_\theta(x)}{2} \log \frac{1+p_\theta(x)}{2} \nonumber\\&&+ \frac{1}{2}\sum_{x'} p_\theta(x') \log p_\theta(x')\nonumber\\
    &=& -\frac{1}{2}\log \frac{1}{2} + \frac{p_\theta(x)}{2}\log \frac{p_\theta(x)}{2} \nonumber\\&&- \frac{1+p_\theta(x)}{2} \log \frac{1+p_\theta(x)}{2}\nonumber\\
    &=& \mathsf{H}_b\left(\frac{1+p_\theta(x)}{2}\right) - \frac{1}{2} \mathsf{H}_b(p_\theta(x)),
\end{eqnarray}
where $\mathsf{H}_b(p) = -p\log p - (1-p)\log (1-p)$ denotes the entropy of a Bernoulli variable. Thus the  $\mathsf{JS}$  divergence  depends  on  the  model  distribution  only  through the probability given by the model to the groundthruth word, $p_\theta(x)$.

\section{Comparative plots of evaluation metrics}\label{sec:plots_metrics}
 
\begin{figure}[t]
    \centering
    \includegraphics[width=1\columnwidth]{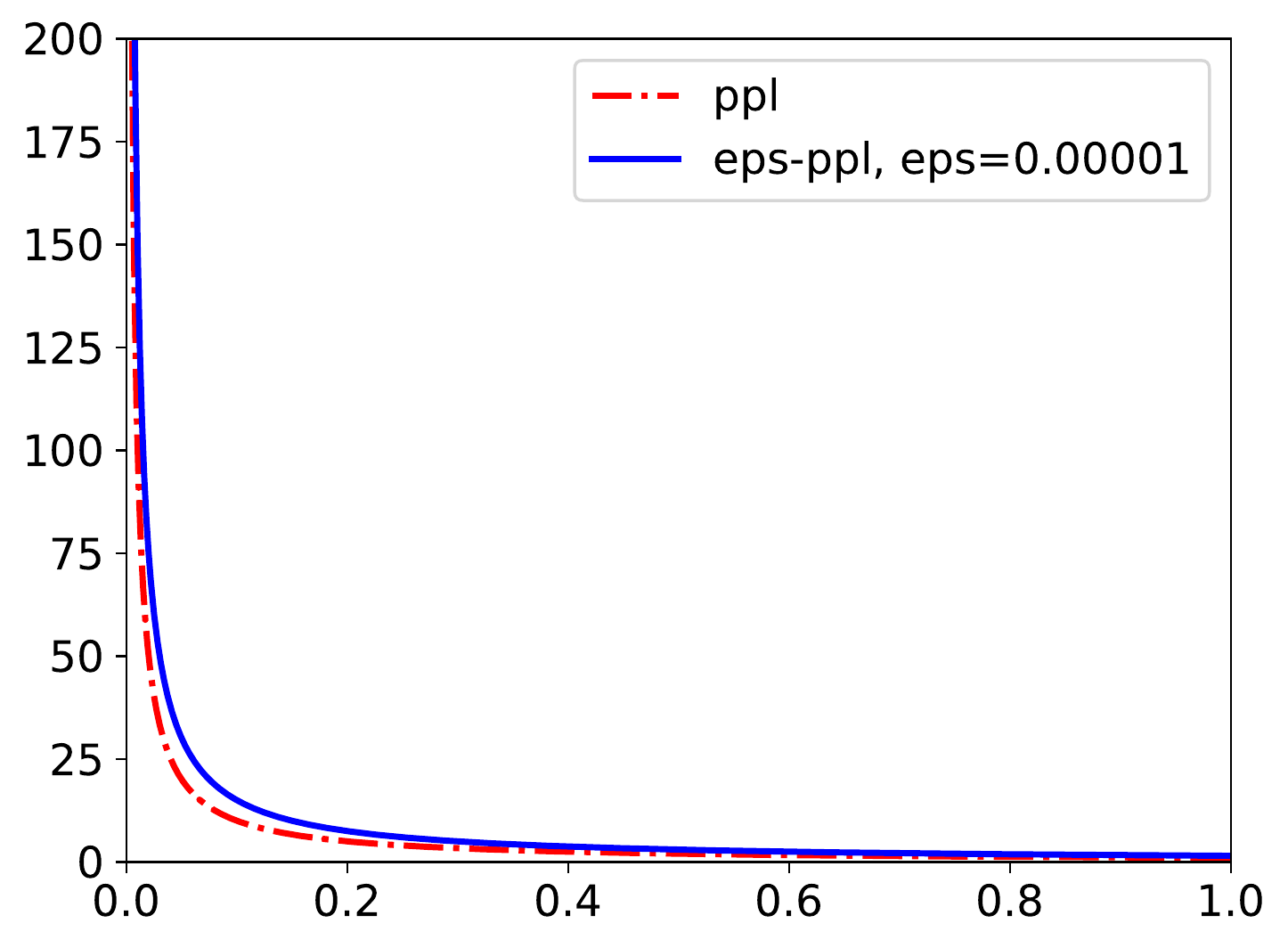}\\
    \includegraphics[width=1\columnwidth]{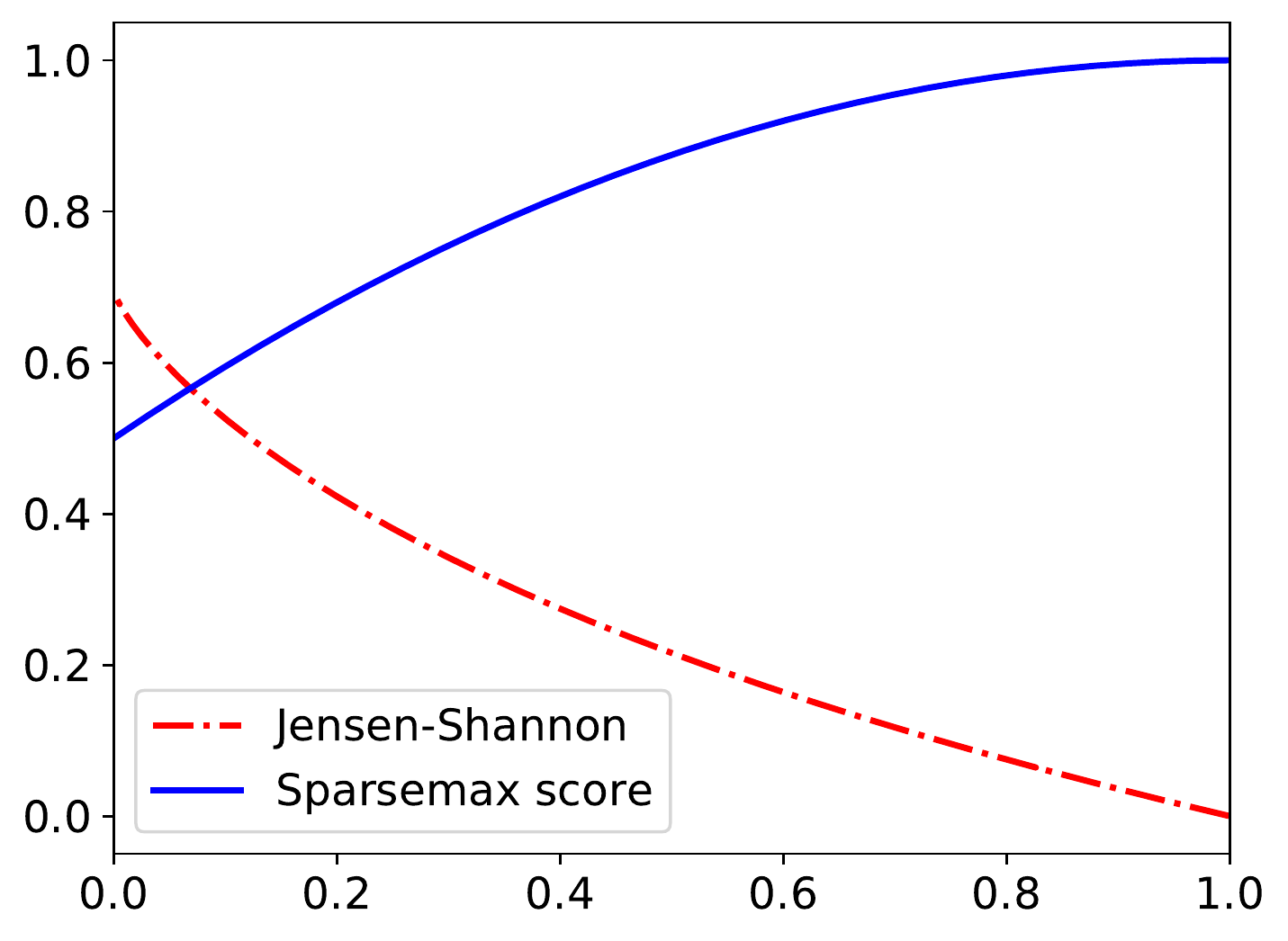}\\
    \caption{Comparative plots of $\epsilon$-perplexity for $\epsilon=0.01$ and $\epsilon=0$ (top), and of sparsemax score and JS divergence (bottom). In both cases, the $x$-axis is $p_\theta(x)$.}
    \label{fig:plot_ppl}
\end{figure}

Figure~\ref{fig:plot_ppl} shows comparative plots of the $\epsilon$-perplexity, sparsemax score, and Jensen-Shannon divergence, for a distribution of the form $\bs{p}_\theta = \left(1-t, \frac{t}{|\mathcal{V}|-1}, \ldots, \frac{t}{|\mathcal{V}|-1}\right)$, varying $t$, with a vocabulary of 50000 words.

\section{Fine-tuning details for language modeling}
\label{finetuning_lm}
The models were fine-tuned for up to 5 epochs for Wikitext-2 and up to 1 for Wikitext-103 and BookCorpus using the Adam optimizer \citep{kingma2014adam}, with a learning
rate of $6.25\times 10^{-5}$, which was linearly decayed to zero over the course of training. We report results of the models that have the highest sparsemax score on the validation set. 
The models fine-tuned with cross entropy and entmax losses were trained on a GPU Nvidia Titan XP, which has $\approx12$ Gb of memory. The model fine-tuned with the unlikelihood training term was trained on a GPU Nvidia Titan RTX, which has $\approx24$ Gb of memory.

\begin{table*}[t]
\vspace{\baselineskip}
\centering \small
\setlength{\tabcolsep}{0.7ex}
\resizebox{\textwidth}{!}{
\begin{tabular}{l@{\hspace{0.5ex}}rrrrr@{\hspace{2.5ex}}rrrrr@{\hspace{2.5ex}}rrrrr}
\toprule
& \multicolumn{5}{c}{WikiText-2} & \multicolumn{5}{c}{WikiText-103}& \multicolumn{5}{c}{BookCorpus}\\
\midrule
     & \textsf{sp} & \textsf{JS} & $\epsilonppl$ & \textsc{rep} & \textsc{wrep} & \textsf{sp} & \textsf{JS} & $\epsilonppl$ & \textsc{rep} & \textsc{wrep} & \textsf{sp} &\textsf{JS} & $\epsilonppl$ & \textsc{rep} & \textsc{wrep} \\
\midrule
Softmax & $.682$ & $.381$ & $\textbf{13.56}$ & $\textbf{.389}$ & $\textbf{.173}$ & $.682$ & $.376$ & $13.26$ & $.342$ & $.162$ & $.691$ & $.360$ & $\textbf{9.56}$ & $.377$ & $.174$\\
Softmax-$\tau$ & $.681$ & $.374$ & $13.62$ & $.403$ & $.178$ & $.681$ & $.358$ & $13.35$ & $.353$ & $.168$ & $.689$ & $.348$ & $9.75$ & $.391$ & $.248$ \\
Greedy & $.484$ & $\textbf{.358}$ & $533.03$ & $.512$ & $.232$ & $.486$ & $\textbf{.357}$ & $523.68$ & $.445$ & $.211$ & $.508$ & $\textbf{.341}$ & $946.03$ & $.456$ & $.198$\\
Top-$k$ & $.680$ & $.368$ & $22.23$ & $.426$ & $.198$ & $.679$ & $.360$ & $22.28$ & $.368$ & $.182$ & $.688$ & $.347$ & $19.55$ & $.398$ & $.193$\\
Nucleus & $.681$ & $.375$ & $15.38$ & $.400$ & $.176$ & $.681$ & $.363$ & $15.65$ & $.352$ & $.167$ & $.690$ & $.348$ & $14.58$ & $.392$ & $.183$\\
Unlikelihood & $.468$ & $.369$ & $635.02$ & $.441$ & $.205$ & $.471$ & $.367$ & $613.61$ & $.411$ & $.196$ & $.492$ & $.352$ & $486.65$ & $.446$ & $.196$ \\
Entmax & $\textbf{.684}$ & $.376$ & $14.69$ & $.397$ & $\textbf{.173}$ & $\textbf{.686}$ & $.362$ & $\textbf{13.25}$ & $\textbf{.341}$ & $\textbf{.160}$ & $\textbf{.699}$ & $.351$ & $9.57$ & $\textbf{.375}$ & $\textbf{.170}$\\
\bottomrule
\end{tabular}
}
\caption{Language model evaluation on WikiText-2, WikiText-103, and BookCorpus validation sets. For all metrics except \textsf{sp}, lower is better.}
\label{results_lm_auto_val}
\end{table*}

\section{Stories' human evaluation details}
\label{human_evaluation_details_stories}

\begin{figure*}[t]
    \centering
    \includegraphics[width=15cm]{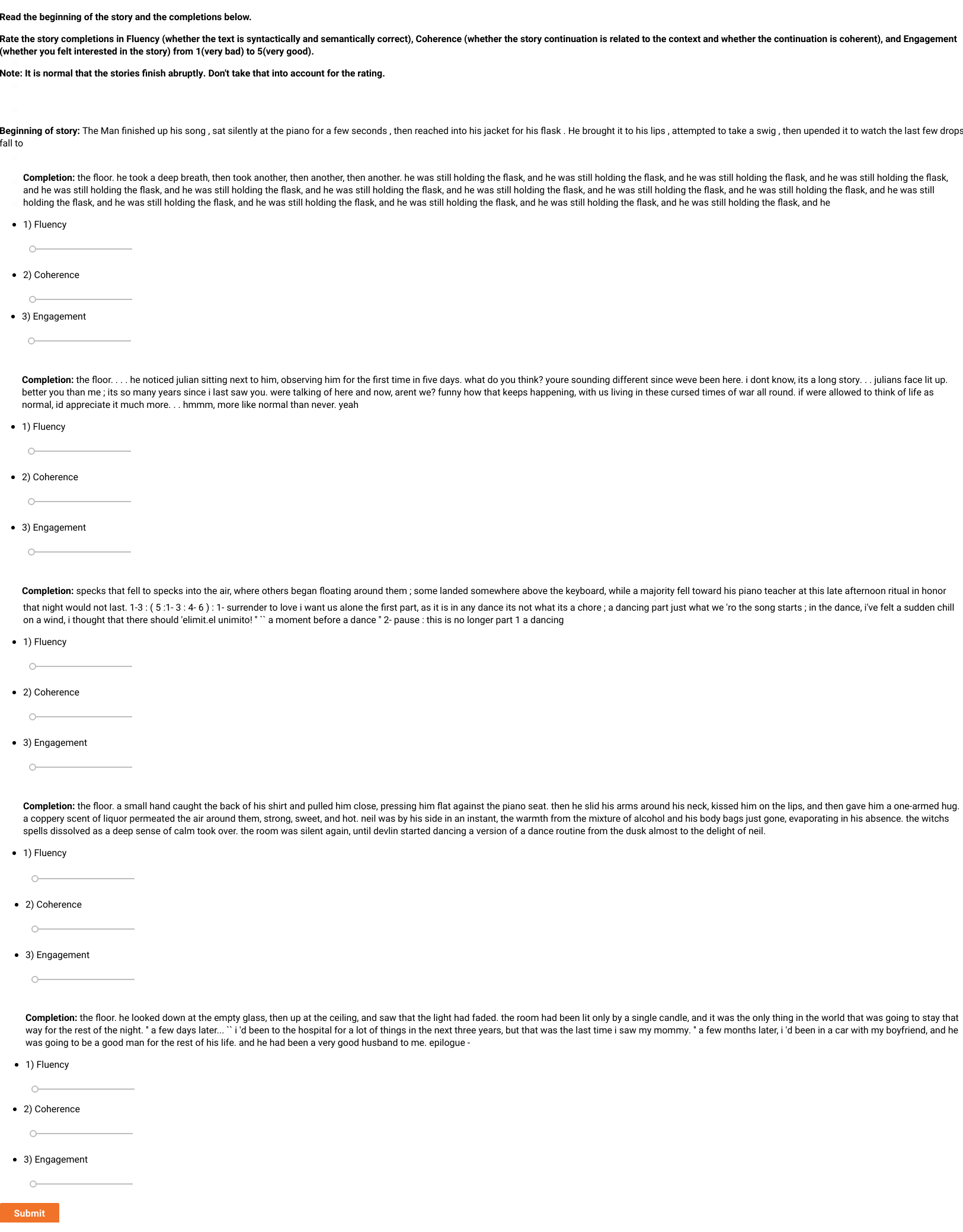}\\
    \caption{Screenshot of human evaluation interface on Amazon Mechanical Turk}
    \label{fig:screenshot_amt}
\end{figure*}

To perform the human evaluation of the stories generated by the different models, we use Amazon Mechanical Turk (a screenshot of the interface is shown in Figure~\ref{fig:screenshot_amt}), and compensate Turkers at a rate of $\$0.7$ per HIT. Pay rate is calculated based on an estimate of the completion time (5.5 minutes) and an hourly wage of $\$7.5$.

To remove poor quality annotations, we perform several controls. We did not consider annotations that were performed in less than 3.5 minutes. Additionally, following \citep{li2019acute}, to filter low quality annotators we showed them annotated examples with contexts from famous novels, the real continuation, story continuations that are not related to the context, and story continuations that are not fluent. If the Turker's annotations differed significantly from the reference rank-wise, all annotations performed by the Turker were excluded.

\section{Fine-tuning details for dialogue generation}
\label{finetuning_dialogue}
We fine-tune the GPT-2 medium model \citep{radford2019language} for a maximum of 3 epochs with a learning rate of $6.25\times 10^{-5}$ that linearly decays to zero over the course of the training. The models were fine-tuned on a GPU Nvidia Titan XP, which has $\approx12$ Gb of memory.

\section{Comparison of models with the Jensen-Shannon divergence}
\label{js_comp_models}

We compared the distributions given by the different decoding methods when generating the sentences of the BookCorpus validation set with the Jensen-Shannon divergence, as described in \S\ref{js}. In Tables \ref{example_js_1} and \ref{example_js_2} we show some of the sentences with higher Jensen-Shannon divergence, as well as the probability given by each model to the ground truth word.

\begin{table*}[ht]
\vspace{\baselineskip}
\centering \small
\setlength{\tabcolsep}{1.5ex}
\begin{tabular}{lcccccccccc}
\toprule
     &  `` & besides & , & i & enjoyed & having & her & with & me\\
\midrule
Softmax & 0.011 & 0.0002 &  0.808 & 0.1479 & 0.0002 & 0.0141 &  0.0228 & 0.0179 & 0.9114   \\
Softmax-t & 0.0131 & 0.0001 & 0.8855 &  0.1864 & 0.0001 & 0.0137 &  0.0211 &  0.0179 &  0.9467  \\
Greedy & 0 & 0 & 1 & 1 & 0 & 0 & 0 & 0 & 1    \\ 
top-$k$ & 0.0439 & 0 & 0.8814 & 0.2543 & 0 & 0 & 0.0311 & 0 & 0.9267          \\
Nucleus  & 0.037 & 0 &  1 & 0.3042 & 0 & 0 & 0 & 0 & 1      \\
Entmax & 0.0159 & 0 & 0.9943 &  0.3311 & 0 & 0.044 & 0.0073 & 0.0185 &  1         \\
\bottomrule
\end{tabular}
\caption{Probabilities given by each model to ``besides, i enjoyed having her with me''}
\label{example_js_1}
\end{table*}

\begin{table*}[ht]
\vspace{\baselineskip}
\centering \small
\setlength{\tabcolsep}{1.5ex}
\begin{tabular}{lccccccccc}
\toprule
     &  i &  miss & my & parents & and & my & friends\\
\midrule
Softmax & 0.0299 & 0.0006 & 0.0294 & 0.0104 & 0.1031 & 0.0611 & 0.0225  \\
Softmax-t & 0.0399 &  0.0005 & 0.0279 &  0.0121 & 0.1067 & 0.0763 & 0.0240\\
Greedy & 0 & 0  &  0 &  0 &  0 & 1 & 0\\ 
top-$k$ & 0.1193 &  0 & 0.0391 & 0 & 0.01303 & 0.1410 & 0  \\
Nucleus  & 0.1005 &  0 &  0 & 0.0289 & 0.1035 &  0.1012 & 0 \\
Entmax & 0.1047 & 0 & 0.0320 & 0.0127 &  0.1453 & 0.1509 & 0.0469  \\
\bottomrule
\end{tabular}
\caption{Probabilities given by each model to  ``i miss my parents and my friends''.}
\label{example_js_2}
\end{table*}

\section{Results of automatic metrics for various values of $K$,$P$, $\alpha$.}
\label{app:sweep}
\begin{figure*}[]
\begin{center}
\includegraphics[width=16cm]{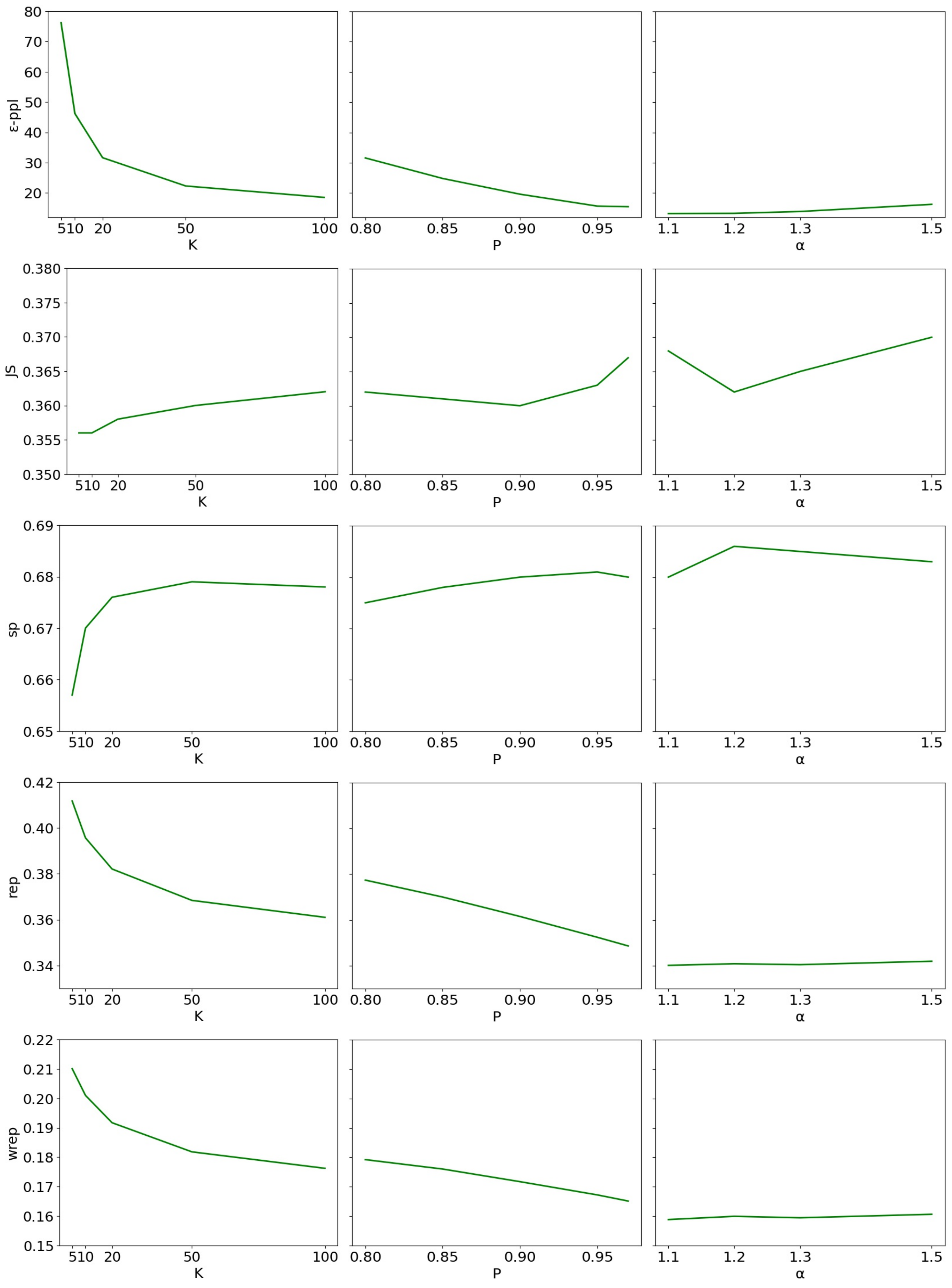}
\end{center}
\caption{Results of automatic metrics on the validation set of WikiText-103 for various values of $K$ for top-$k$ sampling, $P$ for nucleus sampling, and $\alpha$ for entmax. For all metrics except $\mathsf{sp}$, lower is better.}
\label{sweep}
\end{figure*}

In Figure~\ref{sweep} we report the results of $\epsilonppl$, $\mathsf{JS}$, $\mathsf{sp}$, $\mathsf{rep}$, and $\mathsf{wrep}$ metrics on the validation set of WikiText-103 for the models with top-$k$, nucleus, and entmax sampling with various values of $K$,$P$, $\alpha$.

\section{Story completion examples}
\label{app:story_examples}
Examples of story completion with context from the WritingPrompts dataset using the different models fine-tuned on BookCorpus are presented in Table~\ref{examples_stories}.\begin{table*}[ht]
\vspace{\baselineskip}
\centering \small
\setlength{\tabcolsep}{0.5ex}
\begin{tabular}{lp{0.8\textwidth}}
\toprule
\textbf{Context:} &  \textit{``Where am I?'' I asked. I stared at the screens, brushing past my field of vision. They went too fast to watch, but it was clear. I stood on a floating white platform, below me there were flames ...}
\\
\midrule
\textbf{Greedy:}  &... . i could see the flames, but i couldn't see the man. i couldn't see the man who had been my father. i couldn't see the man who had been my father. i couldn't see the man who had been my father. i couldn't see the man who had been my father. i couldn't see the man who had been my father. i couldn't see the man who had been my father. i couldn't see the man who had been my father. \vspace{0.3cm}\\ 
\textbf{Top-$k$:} & ... and a large red door. they went faster and farther off, so close now you almost saw nothing through them as everything around you was just flames; you could taste a burnt smell and you were flying above yourself at speed you never seemed close compared when one thing was the sky or nature or a building full of air around it the sky being what the nature had to look like it had an energy about to burn.   \vspace{0.3cm}\\
\textbf{Nucleus:} & ... and a third being. the creature had clawed its way through the sky, like a giant serpent, its head lifts and back-end leans, it had wings in front and they were on fire, blotched blue and green, emulating the lights in the air around it, like a creature through a screen. i recalled every detail of that moment of events, that once of its kind it had tried to eat me. now i could see the terror in that moment.  \vspace{0.3cm}\\
\textbf{Unlikelihood}& ... and a black smoke. the smoke was rising from my hand, and i could see it was coming from my mouth, and it was the same thing that had killed the woman in the park. ``i don't know how long i can live without you'' i said, and we both knew that the next time we saw each other again, i was going to be a very different person. is the beginning of a new life a man who is the beginning the world \vspace{0.3cm}\\
\textbf{Entmax:} & ... and a demon. the room grew dimming with tiny figures visible as small shadows. we waited. i listened to the tingling below me. a faint noise. the darkness seemed to get thinner. i looked up at her. nothing met my eyes. the girl was familiar, but i couldn't get it. she shuddered. ``no!'' i yelled. her eyes snapped up to me. she screamed, ``no, you are not ready!'' she sprinted towards the door. \vspace{0.15cm}\\
\midrule
\textbf{Context:} &  \textit{The universe was collapsing in on itself , swallowing all that man had once nurtured and loved and explored . As the walls came crashing down and the void beyond it all came flooding in , a man at the center of it all sat serenely before a rift in ...} \\
\midrule
\textbf{Greedy:}  &  ... the universe, a man who had been a man for so long that he had forgotten what it was like to be a man. he was a man who had been a man for so long that he had forgotten what it was like to be a man. he was a man who had been a man for so long that he had forgotten what it was like to be a man. he was a man who had been a man for so long that he had forgotten what it was like to be a man. \vspace{0.3cm}\\
\textbf{Top-$k$:} &  ... a river where time could be no closer, an age ago there stood the very center for an entire generation : that great hour upon timoria before its collapse. it may well also have been that at some early meeting among them the young men at least could give any possible threat a warning, for, one can not blame those boys on this side. they know too well where a great danger rests, or is waiting, somewhere \vspace{0.3cm}\\
\textbf{Nucleus:} & ... the fabric of light that was the void itself.  consuming him whole it will fall. it will swallow us whole. kill us. die. and when it does then it will start over. the vast blackness continued to swallow sebastian, forcing him forward. he twisted, protesting the pain that gathered in his throat as his back was crushed against the solid blackness. he had been doing that all the way down. \vspace{0.3cm}\\
\textbf{Unlikelihood}& ... the fabric of time. it was the first time he had seen a real time machine, a machine with a purpose.'' ``what was it?'' the man who was the future is the beginning of a part whatever the end whatever all yours and whatever in the beginning the deadly end a deadly ( and deadly deadly ) all yours and the first time he had seen the future \vspace{0.3cm}\\
\textbf{Entmax:} & ... the fabric of space and time and whatever truth he had contained since childhood was now ripening to be broken further. the crushing sadness that seized him made him feel like he had just found a complete lack of reality. it was all like a nightmare. it was all a goddamned nightmare. he shook himself and felt his emotions grow equally furious with himself and the destruction. \\
\bottomrule
\end{tabular}
\caption{Examples of story completion using the different methods.}
\label{examples_stories}
\end{table*}

\section{Dialogue simulation examples}
\label{app:sim_examples}
In order to evaluate the different decoding methods in dialogue generation, we simulated a conversation between two agents using the same decoding method. For that, we assigned each agent a randomly selected persona from the PersonaChat dataset and gave a first utterance of the dataset as context.
Examples of dialogue simulations are presented in Tables \ref{examples_dialogue_1}-\ref{examples_dialogue_2} and Tables \ref{examples_dialogue_3}-\ref{examples_dialogue_4}.

\begin{table*}[h!]
\vspace{\baselineskip}
\centering \small
\setlength{\tabcolsep}{0.5ex}
\begin{tabular}{lp{0.7\textwidth}}
\toprule
\textbf{Context: } &  \textit{hello how are you today?}
\\
\midrule
\textbf{Softmax:}  & - i'm great i love to dance to oldies music was is fun\\&
 - very fluent in all the vernacular spoken in pa\\&
 - that's awesome i teach elementary school\\&
 - i love teaching, tap dancing to bob segar, all illustration type stuff\\&
 - that sounds like a lot of fun\\&
 - it is, but this harry potter series just made me feel tingly\\&
 - that's really hurtful to do\\&
 - it can be, but i am working on other things\\&
 - that you are fine i like james patterson\\&
 - very nice of you to comment\\&
 - yes my dog is my son that he chews candy to be funny for my dogs birthday\\&
 - i have no kids i moved to virginia\\&
 - that is crazy did you like the hurricane ?\\&
 - yeah it was so devastating\\&
 - i see\\&
 - its so sad i'm sorry you are stuck in pennywise\\&
 - wow that's very sad\\&
 - it is he?\\&
 - seriously\\&
 - it will be my big dream to leave and go to florida\vspace{0.3cm}\\ 
\textbf{Softmax-$\tau$:}  & - i am doing good, how about yourself?\\&
 - i am good, just getting in from work\\&
 - oh, what do you do?\\&
 - i sell cars, what do you do?\\&
 - i work in a hospital as a nurse.\\&
 - oh that is very nice of you\\&
 - what do you sell cars ?\\&
 - i sell cars as a hobby\\&
 - oh, i see. do you have any hobbies?\\&
 - i like to paint and hang out in my living room.\\&
 - oh, i enjoy listening to music. i love rolling stones !\\&
 - i like rolling stones too do you?\\&
 - yeah, i really like the rolling stones, they're my favorite!\\&
 - cool, meat is amazing. i also like to like painting\\&
 - i think it is for the best. i like to know that i love my job.\\&
 - yes beater and i love meat. i also love to eat meat.\\&
 - i'm sure you are a vegetarian! what do you do for a living?\\&
 - i sell cars, i'm an electrician\\&
 - oh i see. are you a car salesman?\\&
 - i just started selling cars \vspace{0.3cm}\\ 
\textbf{Greedy:}  & - i'm good, how are you?\\&
 - i'm good thanks for asking\\&
 - what do you do for a living?\\&
 - i'm studying to be a dentist\\&
 - oh, i'm a teacher.\\&
 - what city are you from?\\&
 - i live in the city.\\&
 - what do you do for a living?\\&
 - i teach math and english.\\&
 - what city are you from?\\&
 - i live in the city.\\&
 - what city are you from?\vspace{0.3cm}\\ 
\bottomrule
\end{tabular}
\caption{Example of dialogue simulation between two agents using the different decoding methods.}
\label{examples_dialogue_1}
\end{table*}

\begin{table*}[h!]
\vspace{\baselineskip}
\centering \small
\setlength{\tabcolsep}{0.5ex}
\begin{tabular}{lp{0.7\textwidth}}
\toprule
\textbf{Context: } &  \textit{hello how are you today?}
\\
\midrule
\textbf{Top-$k$:}  & - hello! i am well ,yourself ?\\&
 - i am great. i just got off of work\\&
 - i just got a job at a grocery store.\\&
 - oh i am sorry to hear that\\&
 - i'm okay. do you like reading ?\\&
 - yes i love reading and i also love to swim\\&
 - oh that's awesome! i really like reading cupcakes.\\&
 - that sounds delicious. i'm in the south so i like to eat cupcakes\\&
 - i love to make my own cupcakes. i love it!\\&
 - do you have any hobbies?\\&
 - i love reading, but my favorite hobby is painting.\\&
 - that is cool. i've been to canada twice , but i've never been there\\&
 - i love to paint too! what color eyes do you have?\\&
 - mine are blue i'm not sure\\&
 - i have brown hair.\\&
 - i like to make my own cupcakes\\&
 - cool. do you work? i work at walmart and i work at the grocery market.\\&
 - i am actually a reporter for the local news\\&
 - cool! do you like to read?\\&
 - i've a newspaper reporter in the news\vspace{0.3cm}\\ 
\textbf{Nucleus:}  & i'm great, just got home from a long walk with my two dogs.\\&
 - do you play any instruments?\\&
 - not really. what kind of music do you like?\\&
 - i play piano sometimes and listen to metal music\\&
 - oh cool. i'm just trying to write an ebook about animals.\\&
 - i like cats and dogs. do you play any instruments?\\&
 - i do not, i prefer listening to music i like.\\&
 - what kind of music do you like?\\&
 - classical music and video game soundtrack.\\&
 - do you go to concerts?\\&
 - i go with my husband and listen to a lot of different music genres.\\&
 - do you have any kids?\\&
 - nope, how about you?\\&
 - i've a boyfriend. are you married?\\&
 - yes, recently divorced. we are planing on getting a corvette.\\&
 - how exciting! are you close to your husband?\\&
 - i am. i am looking to buy one when i retire.\\&
 - that is awesome! are you married?\\&
 - nope. we were looking to move to ireland when i got my divorce.\\&
 - that's so sad. you should i ask if he's ok?  \vspace{0.3cm}\\ 
\textbf{Entmax:}  & - good, how are you doing?\\&
 - great just watching a movie\\&
 - oh really what movie?\\&
 - it's a story about robots getting into human stuff.\\&
 - wow! well that strikes me as not science fiction at all\\&
 - why do you say that then?\\&
 - because humans are fragile and vulnerable lol\\&
 - yeah they would be\\&
 - very true. what kind of movies u like?\\&
 - i love sci fi and ghost in the shell\\&
 - i actually have never watched that. maybe ghost in the shell\\&
 - it is such a beautiful movie\\&
 - yeah i love stories about lost ones\\&
 - that 's cool\\&
 - my favourite movie though is gamera how about you\\&
 - i love it too, did you see it?\\&
 - yes that was one of my favorite films\\&
 - that's cool. do you work?\\&
 - i am too lazy to try difficult things now\vspace{0.3cm}\\ 
\bottomrule
\end{tabular}
\caption{Example of dialogue simulation between two agents using the different decoding methods.}
\label{examples_dialogue_2}
\end{table*}

\begin{table*}[h!]
\vspace{\baselineskip}
\centering \small
\setlength{\tabcolsep}{0.5ex}
\begin{tabular}{lp{0.7\textwidth}}
\toprule
\textbf{Context: } &  \textit{greetings from the big apple, kim here, how are you?}
\\
\midrule
\textbf{Softmax:}  & - hi! great thanks, how are you ? do you have any hobbies?\\&
 - i ride my black mbu on my nightly walks, and listen to rock music.\\&
 - whats your favorite kind of music?\\&
 - guns and roses is my bedrock, my favorite artist.\\&
 - that's great, my favorite genre is joleg tonne or tontera band.\vspace{0.3cm}\\ 
\textbf{Softmax-$\tau$:}  & - i'm doing great, how are you?\\&
 - i'm good, just playing with my son today.\\&
 - sounds fun, what are you playing with?\\&
 - I'm playing soccer right now, he's 4.\\&
 - do you have any pets?\\&
 - no, just the baby. are you married?\\&
 - no, no i don't have any kids\\&
 - i'm single, but i have been in this profession for a while.\\&
 - interesting, tell me more about yourself.\\&
 - i'm a professional soccer player. what about you ?\\&
 - that's cool, i'm just a training dummy\\&
 - i also play soccer professionally, i love it.\\&
 - i like to play sports too. definitely a great sport\\&
 - i have been to stanford for training, trying to perfect my skills.\\&
 - i bet that's great training\\&
 - its a lot of work!\\&
 - yeah but i like it\\&
 - yeah. i'm actually a soccer mom, we try to keep on top of our kids.\\&
 - do you have kids?\\&
 - i do.\\&
  \vspace{0.3cm}\\ 
\textbf{Greedy:} & - i'm doing well. how are you?\\&
 - i'm good, just finished cooking dinner.\\&
 - what do you like to cook?\\&
 - i love pasta, and i love to cook.\\&
 - i like to make pasta too. i'm a huge fan of pasta.\\&
 - i love to cook, i'm a student, so i'm always busy.\\&
 - what are you studying?\\&
 - i'm studying to be a chef.\\&
 - what city are you from?\\&
 - i'm from cali, what about you?\\&
 - i'm from cali. i'm from the midwest.\\&
 - i'm from the midwest too , but i'm from the midwest.\vspace{0.3cm}\\ 
\bottomrule
\end{tabular}
\caption{Example of dialogue simulation between two agents using the different decoding methods.}
\label{examples_dialogue_3}
\end{table*}

\begin{table*}[h!]
\vspace{\baselineskip}
\centering \small
\setlength{\tabcolsep}{0.5ex}
\begin{tabular}{lp{0.7\textwidth}}
\toprule
\textbf{Context: } &  \textit{greetings from the big apple, kim here, how are you?}
\\
\midrule
\textbf{Top-$k$:}  &  - i'm very good, thank you, i hope you have a great day\\&
 - thanks. have a good day too.\\&
 - same to you, have a good day too!\\&
 - you do the same. have a great day too.\\&
 - the same to you and have a great day also!\\&
 - i do have a very nice day, i love running.\\&
 - that's nice to enjoy, it s a great day here today\\&
 - i am a political junkie.\\&
 - you must be very busy!\\&
 - what do you do for a living?\\&
 - i'm actually a business major.\\&
 - that must be hard to do, do you like politics?\\&
 - not a big fan of politics, i like to go skiing and listen to rock\\&
 - do you play sports or something?\\&
 - i am in school for business\\&
 - oh that is awesome. i used to play football in college, but i hated it!\\&
 - oh that's why i play football!!\\&
 - wow. what is your name?\\&
 - my name is jed. i am from russia. how about you?\\&
 - i played in ny and played at boston.\\&
 \vspace{0.3cm}\\ 
\textbf{Nucleus:}  &  - hi, i'm fantastic, how are you?\\&
 - great. making some barbies for our barbie girls.\\&
 - awesome, do you have any hobbies?\\&
 - i really enjoy barbies. i live a long time away.\\&
 - i am looking forward to going to school in the fall.\\&
 - school is cool. do you sing? i love barbie songs.\\&
 - i've a pet pean i sing while eating tacos.\\&
 - barbie! that's cute.\\&
 - oh yea, i have a small heart lol\\&
 - are you looking forward to winter? i would like to live in the big apple.\\&
 - definitely\\&
 - winter is fun! no thanks, me neither.\\&
 - you must really like summer, too lol\\&
 - thanks. it is too cold to sing in the winter. its just not for me.\\&
 - do you like tacos?\\&
 - oh my gosh. they're good. my favorite is barbie. lol\\&
 - i know, but spaghetti is my fave lol\\&
 - haha. lol\\&
 - oh , i do enjoy spaghetti , for my college graduation i was in last year of high school\\&
 - how are you ?  \vspace{0.3cm}\\ 
\textbf{Entmax:}  &  - good. i just rang someone on the other side.\\&
 - good, what did you get excited about today?\\&
 - i love paris. how was your day?\\&
 - its been crazy, i'm always feeling excited! lol\\&
 - what is your favorite thing to do?\\&
 - oh i love going hiking in the mountains, myself.\\&
 - that sound amazing. i like travelling.\\&
 - love trips, but i cannot stand staying in one place all day. lol\\&
 - do you have any hobbies ? i always want to find cool new things.\\&
 - i really like going out and nature itself, i prefer hiking\\&
 - yes, exploring parks and all that jazz when i can.\\&
 - that is awesome fun, whats your fav color?\\&
 - i love grey. roses and the mountains signify my youth.\\&
 - mine is blue, it makes me think of blueberries though\\&
 - grey denotes youth well or openness and transparency. love the kale chips.\\&
 - mmm i love chocolate . lol\\&
 - oh i am sold on chocolate. eating it off the cob\\&
 - haha \\&
 - i miss the crazy curly hair fries and crackers . haha \vspace{0.3cm}\\ 
\bottomrule
\end{tabular}
\caption{Example of dialogue simulation between two agents using the different decoding methods.}
\label{examples_dialogue_4}
\end{table*}

\end{document}